\documentclass[10pt, lettersize,journal]{IEEEtran}
\usepackage{amsmath,amsfonts}
\usepackage{algorithmic}
\usepackage{array}
\usepackage[caption=false,font=normalsize,labelfont=rm,textfont=rm]{subfig}
\usepackage{textcomp}
\usepackage{stfloats}
\usepackage{url}
\usepackage{verbatim}
\usepackage{graphicx}
\usepackage[utf8]{inputenc}
\usepackage[ruled,vlined]{algorithm2e}
\def\BibTeX{{\rm B\kern-.05em{\sc i\kern-.025em b}\kern-.08em
		T\kern-.1667em\lower.7ex\hbox{E}\kern-.125emX}}
\usepackage{balance}
\usepackage{hyperref}
\hypersetup{
	colorlinks = true, 
	linkcolor = blue,
	urlcolor = blue,
	citecolor = blue
}
\usepackage[switch]{lineno}

\begin{document}
		%\pagewiselinenumbers
		% 设置行号在左侧
		
		%\pagewiselinenumbers
		\captionsetup{font={small}}
		\title{Semantic-Aware Spectrum Sharing in Internet of Vehicles Based on Deep Reinforcement Learning}

		\author{Zhiyu Shao, Qiong Wu,~\IEEEmembership{Senior Member,~IEEE}, Pingyi Fan,~\IEEEmembership{Senior Member,~IEEE},
		\\Nan Cheng,~\IEEEmembership{Senior Member,~IEEE}, Wen Chen,~\IEEEmembership{Senior Member,~IEEE}, 
		\\Jiangzhou Wang,~\IEEEmembership{Fellow,~IEEE}, and Khaled B. Letaief,~\IEEEmembership{Fellow,~IEEE}
			% <-this % stops a space
			\thanks{This work was supported in part by the National Natural Science Foundation of China under Grant No. 61701197, in part by the National Key Research and Development Program of China under Grant No.2021YFA1000500(4), in part by the National key project 2020YFB1807700, NSFC 62071296, Shanghai Kewei 22JC1404000, in part by the Research Grants Council under the Areas of Excellence scheme grant AoE/E-601/22-R, in part by the 111 Project under Grant No. B12018. (Corresponding authors: Qiong Wu.)
				
				Zhiyu Shao, Qiong Wu are with the School of Internet of Things Engineering, Jiangnan University, Wuxi 214122, China (e-mail: zhiyushao@stu.jiangnan.edu.cn, qiongwu@jiangnan.edu.cn)
				
				Pingyi Fan is with the Department of Electronic Engineering, Beijing National Research Center for Information Science and Technology, Tsinghua University, Beijing 100084, China (e-mail: fpy@tsinghua.edu.cn).
				
				Nan Cheng is with the State Key Lab. of ISN and School of Telecommunications Engineering, Xidian University, Xi’an 710071, China (e-mail: dr.nan.cheng@ieee.org).
				
				Wen Chen is with the Department of Electronic Engineering, Shanghai Jiao Tong University, Shanghai 200240, China (e-mail: wenchen@sjtu.edu.cn).
				
				Jiangzhou Wang is with the School of Engineering, University of Kent,
				CT2 7NT Canterbury, U.K. (email: j.z.wang@kent.ac.uk).
				
				K. B. Letaief is with the Department of Electrical and Computer Engineering, Hong Kong University of Science and Technology (HKUST), Hong Kong (e-mail:eekhaled@ust.hk).

			}% <-this % stops a space
			% <-this % stops a space
		}

		%% The paper headers
		%\markboth{IEEE Transactions on Vehicular Technology,~Vol.~XX, No.~XX, XXX~2015}%

		\maketitle
		
		\begin{abstract}
			This paper investigates semantic communication in high-speed mobile Internet of Vehicles (IoV), focusing on spectrum sharing between vehicle-to-vehicle (V2V) and vehicle-to-infrastructure (V2I) communications. We propose a semantic-aware spectrum sharing (SSS) algorithm using deep reinforcement learning (DRL) with a soft actor-critic (SAC) approach. 
			We start with semantic information extraction, redefining metrics for V2V and V2I spectrum sharing in IoV environments, introducing high-speed semantic spectrum efficiency (HSSE) and semantic transmission rate (HSR).
			We then apply SAC algorithm to optimize decisions V2V and V2I spectrum sharing decisions on semantic information. 
			This optimization aims to maximize HSSE and enhance the success rate of effective semantic information transmission (SRS), including determining the optimal V2V and V2I sharing strategies, transmission power, and the length of transmitted semantic symbols.
			Experimental results show that the SSS algorithm outperforms other baseline algorithms, including other traditional-communication-based spectrum sharing algorithms and spectrum sharing algorithm using other reinforcement learning approaches. 
			The SSS algorithm exhibits a 15\% increase in HSSE and approximately a 7\% increase in SRS.
			
		\end{abstract}
		
		
		\IEEEpeerreviewmaketitle
		
		\section{Introduction}
		\subsection{Background}
		\IEEEPARstart{I}{n} recent years, researchers have delved into Internet of vehicles (IoV) technology \cite{SGR5, another1}.
		As IoV gains attention, cellular vehicle-to-everything (C-V2X) including communications between vehicles and vehicles (V2V) as well as between vehicles and infrastructure (V2I) become crucial for supporting intelligent transportation systems (ITS) \cite{SGR6}. 
		C-V2X enhances security by leveraging cellular network encryption and authentication protocols, reducing side-channel attack risks and ensure secure data transmission with the rise of security issues \cite{shengairen1}.
		However, the success of IoV depends not only on the rapid development of technology but also on effective spectrum sharing \cite{SGR3, XIN1}. 
		
		The scarcity of spectrum resources has triggered spectrum sharing issues in IoV systems \cite{RNnew}, where V2V communications need to share spectrum resources with V2I communications \cite{RN51}.
		The available radio spectrum is limited and must be divided among various services and users. With the rapid increase in the number of connected vehicles, the competition for this finite resource intensifies \cite{SGR1, SGR2}.
		Vehicles engaged in both V2V and V2I communications simultaneously within the same limited spectrum bandwidth, they need to share the same spectrum resources \cite{XIN5}. This can result in conflicts due to competition for resources as multiple vehicles vie for the same spectrum, potentially causing interference and degradation in communication performance \cite{XIN6}. 
		Additionally, as vehicles move in and out of communication ranges, they may disrupt ongoing communications or access the same spectrum resources concurrently, further exacerbating the issue and impacting the reliability and efficiency of IoV systems. 
		
		Network traffic has become increasingly critical in IoV due to the diverse applications and explosive growth of content, which challenge traditional communication methods for spectrum sharing \cite{RNlala, XIN2}. Vehicles frequently change positions and communication links, leading to fluctuating network loads. Additionally, V2V and V2I applications require timely and accurate data exchange. Network traffic congestion can result in increased latency and message loss.  
		These issues highlight that traditional communication methods, constrained by the Shannon limit (the theoretical maximum rate of information transmission) \cite{RN62}, are inadequate for the IoV environment.
		As the exchange of information between vehicles becomes increasingly rich and complex, especially in the high-speed mobile IoV environment, which surpasses the capabilities of communication methods based solely on bits \cite{RN73}.
		The high latency, limited bandwidth, and interference in vehicular networks can severely impact the communication system, thereby affecting the performance of IoV systems \cite{XIN3}.
		
		The challenge of spectrum scarcity and high network traffic is critical in IoV environments due to the large number of vehicles competing for limited spectrum resources.
		Semantic communication improves efficiency by focusing on contextually relevant and essential information, which minimizes unnecessary data transmission and optimizes spectrum usage \cite{RN78}.
		This approach is particularly beneficial in dynamic vehicular networks, where conditions such as vehicle speed and traffic density change rapidly. By understanding and prioritizing the content and context of the data, semantic communication allows for real-time adjustments in resource allocation, reducing both spectrum occupancy and network traffic \cite{RN80}.
		Additionally, traditional spectrum sharing often wastes bandwidth, while semantic communication reduces spectrum occupancy by focusing on essential data. Finally, by understanding data content and context, semantic communication enhances decision-making in spectrum-sharing algorithms through dynamic transmission power adjustments, optimal link selection, and scheduling based on data relevance and urgency \cite{SGR4}.
		This emerging communication paradigm is considered a potential solution to reduce network traffic and alleviate spectrum scarcity, especially in the context of spectrum sharing in V2V and V2I communication.
		Semantic information enhances security by ensuring data is only meaningful to authorized users who share the same knowledge base. This approach reduces the risk of interception and tampering, as encrypted semantic data cannot be easily interpreted without the correct context.
		
		However, in the high-speed mobile IoV environment, semantic communication still faces specific challenges. 
		First, defining and extracting semantic information from various data streams is complex. Second, traditional vehicle communication relies on bit-based methods derived from statistical knowledge rather than semantic data, necessitating a redefinition of V2V and V2I spectrum sharing for semantic communication. Finally, vehicles must make decisions based on semantic data for spectrum resource allocation. The complexity and richness of semantic information, combined with additional decision variables and nonlinear relationships, render traditional optimization methods inadequate. Therefore, a deep reinforcement learning (DRL) algorithm is needed to effectively leverage semantic information \cite{RN83}. 
		Among DRL methods, Deep Q-Network (DQN) is limited to discrete actions, making it unsuitable for IoV scenarios. While deep deterministic policy gradient (DDPG) can handle continuous actions, it lacks good exploration and stability. The Soft Actor-Critic (SAC) approach not only handles continuous action spaces effectively but also improves sample efficiency and training stability through entropy regularization.
		Additionally, SAC's off-policy nature enables efficient learning by reusing past experiences, which is beneficial in dynamic IoV scenarios.
	
		The objective of this paper is to explore the application of semantic communication in IoV, particularly in addressing spectrum sharing issues. 
		We design a semantic-aware spectrum sharing algorithm (SSS) based on the DRL SAC approach to maximize high-speed semantic spectrum efficiency (HSSE) of V2I and enhance the success rate of effective semantic information transmission (SRS) of V2V links\footnote{The source code has been released at: https://github.com/qiongwu86/Semantic-Aware-Spectrum-Sharing-in-Internet-of-Vehicles-Based-on-Deep-Reinforcement-Learning}.
		We will investigate the extraction of semantic information, redefine spectrum sharing, and optimize decision-making based on semantic data using DRL to enhance the performance of V2V and V2I communications. 
		% By addressing these challenges, our aim is to drive the development of IoV systems, achieving more efficient and reliable ITS.
		\subsection{Related Work and Motivation}
		In this section, we will review the problems of relevant research on the problems of V2V and V2I spectrum sharing in the IoV environment. 
		There have been studies that address the problem using traditional methods.
		In \cite{RN6}, Liang \emph{et al.} proposed a low-complexity algorithm that periodically reports the channel state information (CSI) of vehicle links, addressing spectrum and power allocation issues in highly mobile vehicular networks. The algorithm maximizes V2I link throughput while ensuring the reliability of V2V links.  
		In \cite{RN42}, a graph partitioning and matching approach was then introduced to reduce signaling overhead in vehicular communication by utilizing slow-fading statistical channel information, thus maximizing overall V2I capacity.
		In \cite{RN44}, Guo \emph{et al.} decomposed the V2X resource allocation problem into pure power allocation and pure spectrum allocation subproblems to achieve a globally optimal solution.
		In \cite{RN48}, Liu \emph{et al.} proposed a communication scheme based on a device-to-device (D2D) enabled vehicular communication (D2D-V) distributed robust power control algorithm. This approach, using successive convex approximation and Bernstein approximation of probability functions, addressed cellular user interference constraints in highly mobile environments, optimizing the performance of D2D-V systems.
		
		The aforementioned studies employed conventional methods, whereas \cite{RN49, RN47, RN23, RN61, RN67} applied DRL methods.
		In \cite{RN49}, Ye \emph{et al.} proposed a decentralized resource allocation mechanism for V2V communication based on deep Q-network (DQN) DRL. This approach minimizes interference with V2I communication while meeting strict latency constraints.
		In \cite{RN47}, Liang \emph{et al.} presented a spectrum sharing method for vehicular networks based on multi-agent reinforcement learning with double deep Q-networks (DDQN), enhancing V2I link capacity and V2V link data transfer rates.
		%In \cite{RN19}, Gyawali \emph{et al.} combined graph and deep reinforcement learning to improve the overall throughput of V2X communication through centralized channel allocation and distributed power control.
		In \cite{RN23}, Xu \emph{et al.} proposed a multi-agent DRL using non-orthogonal multiple access technology to address spectrum sharing challenges in vehicular communication. This method ensures the total rate of V2I communication and satisfies strict latency and reliability constraints for V2V communication in highly mobile environments.
		In \cite{RN61}, Yuan \emph{et al.} introduced a meta-learning-based DRL algorithm to enhance resource allocation strategy adaptability in dynamic environments. 
		In \cite{RN67}, Mafuta \emph{et al.} accelerated the learning process of deep neural networks building on a multi-agent DDQN approach through transmission mode selection and binary weight algorithms, reducing computational complexity.
		
		However, whether using traditional methods or reinforcement learning approaches to address spectrum resource allocation problems in the IoV environment, data transmission utilizes conventional bits, rather than the characteristics of semantic communication. These methods often focus more on the quantity and transmission efficiency of data rather than the semantic meaning of data.
		
		Recent research has begun to address the concept of semantic communication, primarily focusing on image, text, and audio data. In the field of image transmission, several studies have proposed semantic transmission methods for images in \cite{RN43, RN30, RN60, RN66}. 
		In \cite{RN43}, Bourtsoulatze \emph{et al.} utilized deep learning-driven joint source-signal and channel coding techniques, treating image pixel values as semantic information to efficiently transmit images in low signal-to-noise ratio and noisy environments.  
		In \cite{RN30}, Huang \emph{et al.}, without considering the channel environment and joint channel-source coding in complex scenarios, primarily employed generative adversarial networks (GANs) to achieve a high-performance semantic image compression scheme. 
		In \cite{RN60}, Yang \emph{et al.} designed a rate-balancing policy network for different channel conditions and image content to achieve a balance between rate and signal quality, utilizing gumbel-softmax trick for differentiability. 
		In \cite{RN66}, Huang \emph{et al.} proposed a reinforcement learning-based adaptive semantic coding (RL-ASC) method capable of encoding images beyond the pixel level.
		
		In the text domain, semantic transmission methods were proposed in \cite{RN33, RN58, RN56}.
		In \cite{RN33}, Xie \emph{et al.} used deep learning to measure sentence similarity as a metric for semantic communication performance, enhancing the robustness of communication systems in low signal-to-noise ratio scenarios.  
		The existing semantic communication system was then lightweighted in \cite{RN58}, incorporating a channel estimation scheme based on the least squares method and a neural network solution for image denoising (ADNET). 
		In \cite{RN56}, Wang \emph{et al.} employed a knowledge graph to model semantic information, optimizing semantic similarity through a proximal policy optimization (PPO) approach with attention networks for resource block allocation.
		
		In the audio domain, some studies proposed semantic transmission methods for audio in \cite{RN57, RN65}. 
		In \cite{RN57}, Weng \emph{et al.} introduced an attention mechanism to assign higher weights to crucial speech information, improving transmission efficiency.  
		In \cite{RN65}, Han \emph{et al.} used a soft alignment module and redundancy removal module based on deep learning to extract only textual semantic features while eliminating semantically redundant content, thereby enhancing transmission efficiency.
		
		For video semantic information research, Tung \emph{et al.} introduced a video semantic transmission method in \cite{RN55}, presenting an end-to-end, source-channel joint coding video transmission mechanism. 
		It directly maps video signals to channel symbols, using reinforcement learning to allocate network bandwidth to maximize overall video visual quality.  
		In \cite{RN64}, Wang \emph{et al.} utilized nonlinear transformations and a conditional coding architecture to adaptively extract semantic features between different video frames.
		
		Despite various methods for extracting semantic information, existing research has not yet addressed the spectrum sharing problem based on semantic information in high-speed mobile IoV scenarios as far as we know. This motivates our proposal of a DRL-based semantic-aware spectrum sharing scheme for high-speed mobile IoV environments.
		\subsection{Contributions}
		In this work, we focus on designing a semantic-aware spectrum sharing algorithm based on DRL, intended for application in high-speed mobile IoV environments. The primary contributions of our work can be summarized as follows:
		\begin{itemize}
			\item[1)] We introduce a novel approach to transform conventional binary data into semantic information, making it suitable for semantic communication. 
			We introduce metrics, high-speed semantic spectrum efficiency (HSSE) and semantic transmission rate (HSR), tailored for high-speed mobile IoV spectrum sharing, replacing traditional metrics of transmission rate (HR) and spectrum efficiency (HSE).
			\item[2)] We formulate a joint optimization problem to maximize HSSE and SRS. This is the first to  introduce semantic communication addressing communication challenges in high-speed mobile IoV environments. 
			\item[3)] We reformulate the optimization problem as a markov decision process (MDP) and utilize a DRL approach, specifically the SAC, for handling continuous variables.
			Based on this, we design semantic-aware spectrum sharing algorithm (SSS).
			\item[4)] Simulation results show our SSS algorithm outperforms traditional communication-based spectrum sharing methods and other reinforcement learning approaches in terms of HSSE and SRS.
		\end{itemize}
		
		The rest of this paper is organized as follows: Section II introduces the system model and the formulated problem of maximizing HSSE and SRS; Sections III and IV present the proposed SSS algorithm; Section V provides and discusses simulation results and Section VI concludes the paper.

		\section{System Model}\label{system}
\begin{figure}[htbp]
			\centering
			\includegraphics[width=0.8\linewidth, scale=1.00]{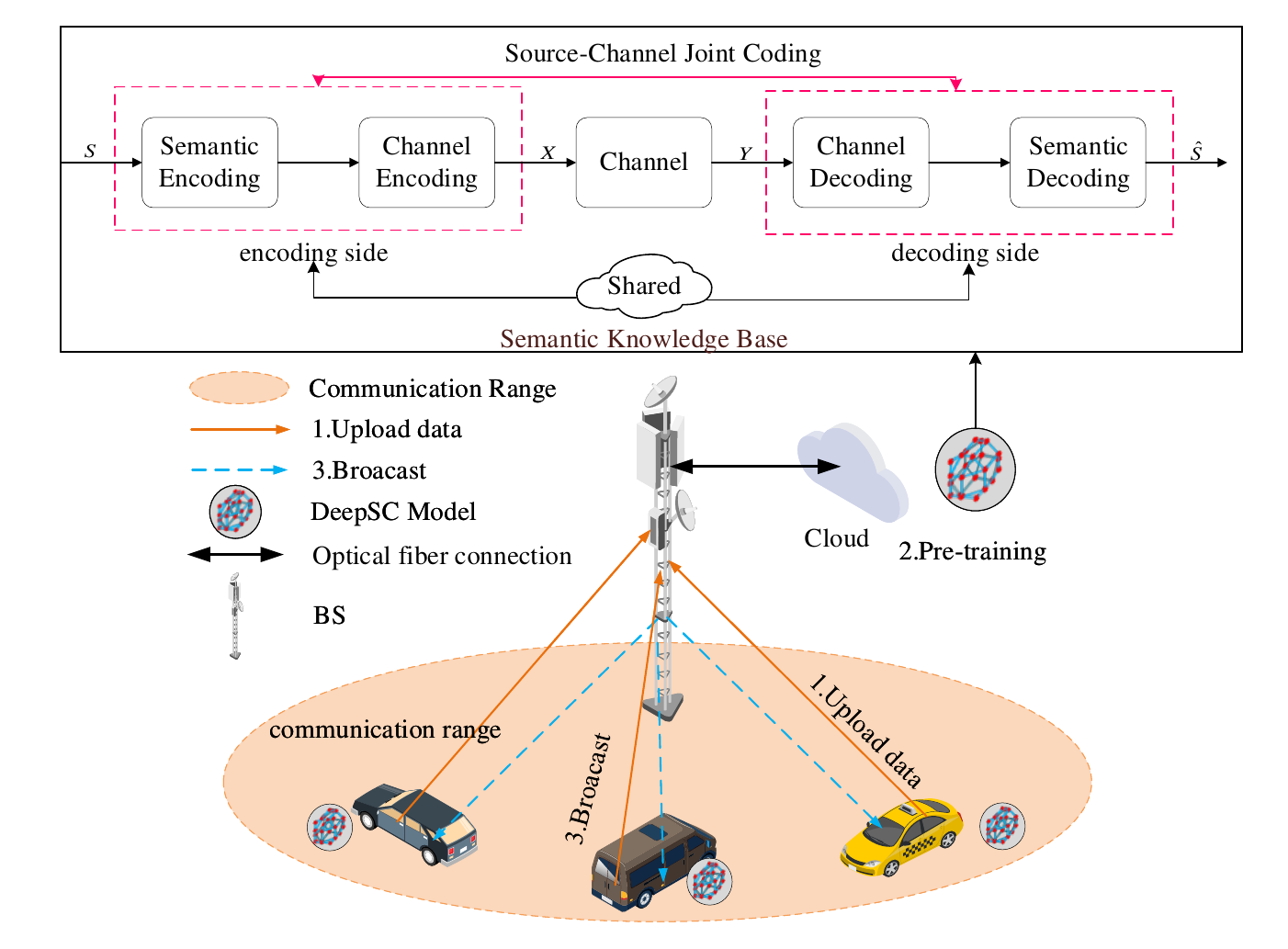}
			\vspace{-0.3cm}
			\caption{Semantic pre-training framework}
			\label{fig2}
		\end{figure}

		As shown in Fig. \ref{fig2}, the DeepSC model enhances communication efficiency by focusing on the meaning of the transmitted data rather than just the raw data itself \cite{RN33}. The model includes an encoding part (semantic and channel encoding), a decoding part (semantic and channel decoding), and a wireless channel, all sharing a common semantic knowledge base.
		Initially, each vehicle collects raw textual data over a wireless channel and transmits it to the base station (BS) within its range. The BS then forwards this data to a central cloud server via a high-speed optical fiber connection, ensuring minimal latency and negligible data loss. The cloud server preprocesses the data and uses it to pre-train the DeepSC model. After pre-training, the cloud server sends the DeepSC model back to the BS, which then broadcasts it to all vehicles, ensuring they have the latest model for local inference. Vehicles use the model to extract semantic information from the textual data they generate and can upload new data or model parameters back to the BS, which forwards them to the cloud server for further training and updates. This feedback loop keeps the model robust and adaptable to changing network conditions.
			
\begin{figure}[htbp]
			\centering
			\includegraphics[width=0.8\linewidth, scale=1.00]{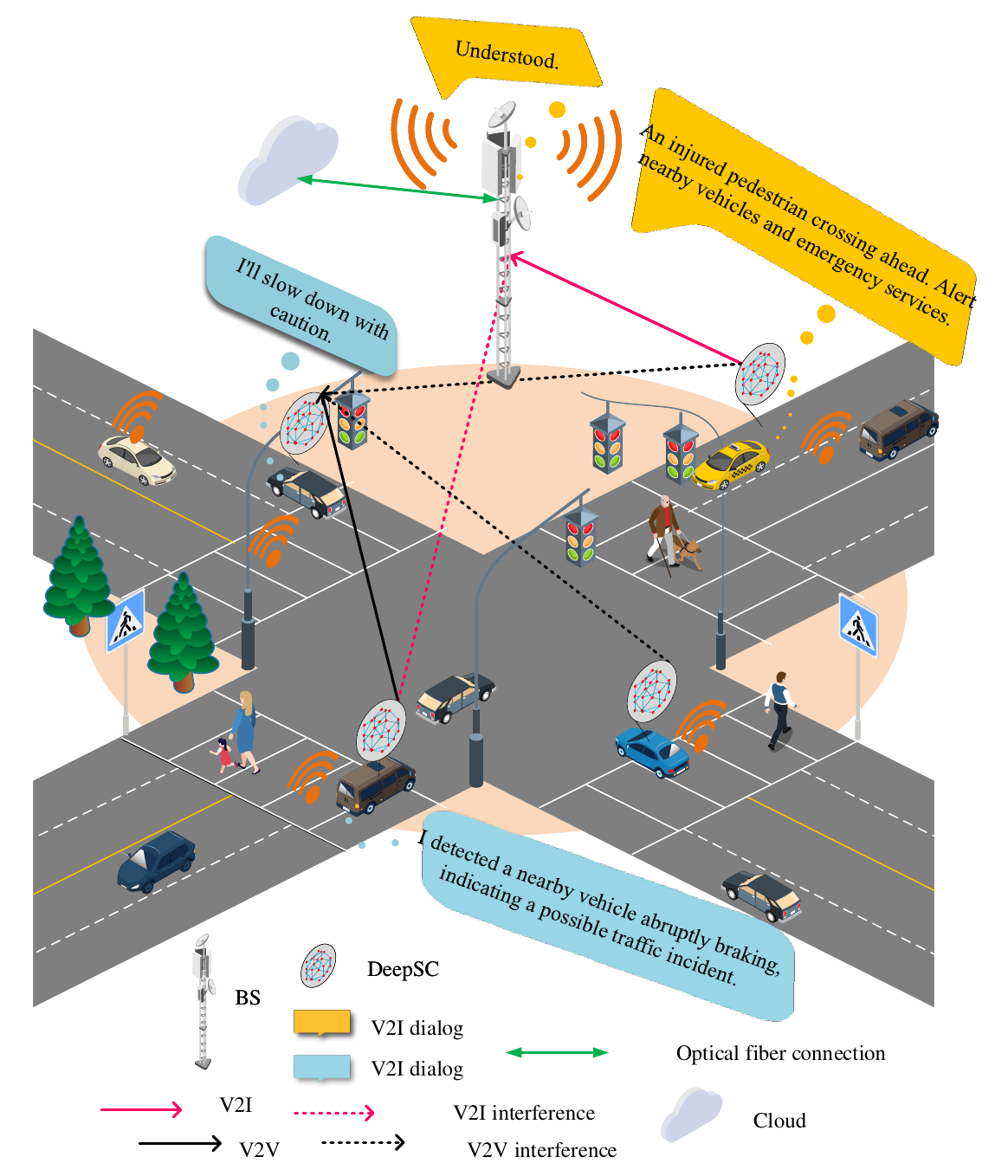}
			\vspace{-0.3cm}
			\caption{The system model}
			\label{fig1}
		\end{figure}

		As shown in Fig. \ref{fig1}, we consider a cellular-based vehicular communication network in urban areas, where $n$ vehicles are driving in the communication range of a BS. The BS is deployed in the center of the intersection. 
		Both BS and vehicles are equipped with DeepSC model.
		The initial positions of vehicles are determined by a spatial Poisson process, and the initial driving direction of each vehicle are set randomly. Each vehicle keeps driving with the initial direction and maintains a constant speed of $v$ km/s. When a vehicle arrives at the intersection, it has a ${p_l}$\% chance to the left direction, ${p_r}$\% chance to the right direction and ${p_s}$\% chance to the straight direction, where ${p_l}\%  + {p_r}\%  + {p_s}\%  = 1$. For keeping the number of vehicles to be the same in the considered scenarios, we also assume that when a vehicle crosses the environment boundary, its position is adjusted to the opposite edge. 
			
		We consider there exists $W$ V2I links and $Q$ V2V links in the system, where the interferences exists among different links.
		As shown in Fig. \ref{fig1}, solid black and red lines represent V2V and V2I communication, respectively, while corresponding dashed lines indicate its related interference. When a vehicle generates a textual dialog for transmission, such as 'I detected a nearby vehicle abruptly braking, indicating a possible traffic incident', it converses the text data into the semantic information based on the DeepSC model, and transmits the semantic information to another vehicle, which could analyze the received semantic information and responds with one appropriate dialog 'I'll slow down with caution'. 

		In the following subsections, we will introduce how DeepSC transceivers extract, transmit, and recover semantic information during the formal process of spectrum sharing for V2V and V2I communication. 	
		Furthermore, interferences in the wireless channel for V2V and V2I communication will be discussed. 
		Finally, communication metrics adapted for semantic communication systems will be proposed.
		\vspace{-0.3cm}
		\subsection{DeepSC Transceivers}
		The transmitting vehicle generates a sentence $S\left[ w \right]$ with ${l}$ words in the $w$-th V2I link as shown in Fig. \ref{fig1}, where $S\left[ w \right] = \left[ {{s_1}\left[ w \right],{s_2}\left[ w \right], \ldots ,{s_l}\left[ w \right]} \right]$.
		 
		\subsubsection{Transmitter Model}
		All the sentences across all links are  $S\left[ w \right]$.
		The sentences are then input into the encoding part of the DeepSC transmitter in the vehicle in order to extract semantic information $X\left[ w \right]$ from $S\left[ w \right]$. 
		The semantic encoding part utilizes advanced NLP technique (Transformer-based encoding) to extract semantic information. This information is then encoded into a suitable format for transmission by the channel encoding part.
	    The formula is given by:
		\begin{equation}\label{eq1}
			\vspace*{-0.2\baselineskip}
			X\left[ w \right] = c{h_\beta }\left( {s{e_\alpha }{\rm{ }}\left( {S\left[ w \right]} \right)} \right),
		\end{equation}
		where $s{e_\alpha }{\rm{ }}\left( \cdot \right)$ and $c{h_\beta }\left( \cdot \right)$ are the semantic and channel encoder networks with parameters $\alpha$ and $\beta$ respectively.
		The semantic symbol vector is $X\left[ w \right] = {x_1}\left[ w \right],{x_2}\left[ w \right], \ldots ,{x_l}\left[ w \right], \ldots {x_{u_ql}}\left[ w \right]$, where $X\left[ w \right] \in {R^{{u_q}l}}$ and ${u_q}$ represents the average number of semantic symbols used for each word.
		The vector dimension is determined by the semantic and channel encoders' neural network.
		The value of $l$ is fixed based on the European parliament dataset \cite{RN91}, which contains predefined English sentences used for training, thus $l$ is known.
		Consequently, ${u_q}$ can be computed by dividing the overall vector of ${X\left[ w \right]}$ by the corresponding sentence length $l$ from the dataset.
		\subsubsection{Transmission Model}
		We focus on Mode 4 defined in the cellular V2X architecture \cite{RN85}, but the transmission is no longer in bits but in semantic information.
		We suppose each of $W$ V2I uplinks has been assigned a specific frequency sub-band and uses a fixed transmission power. Specifically, the $w$-th V2I link uses the $w$-th sub-band.
		
		For the $w$-th V2I link, its channel gain on the $w$-th sub-band is denoted as ${h_{w,B}}\left[ w \right]$. 
		The channel gain for the $q$-th V2V link, utilizing the $w$-th sub-band for V2V communication, is represented by ${h_q}\left[ w \right]$, which is represented as:
		\begin{equation}\label{eq40}
			\vspace*{-0.2\baselineskip}
			{h_q}\left[ w \right] = s{s_q}\left[ w \right] \times l{s_q},
		\end{equation} 
		 $s{s_q}\left[ w \right]$ is frequency-dependent small-scale fading due to multi-path propagation and interference, which follows an exponential distribution with $s{s_q}\left[ w \right] \sim {\rm{Exp(1)}}$. 
		 $l{s_q}$ is frequency-independent large-scale fading due to path loss and shadowing effects. 
		
		However, there is interference on both V2I and V2V links.
		For V2I communication, the interference ${h_{qB}}\left[ w \right]$ is channel gain originating from the transmitter of the $q$-th V2V link to the BS on the $w$-th sub-band. 
		For V2V communication, two types of interference arise. 
		The first is ${h_{q'q}}\left[ w \right]$ originating from the transmitter of the $q'$-th V2V link to the receiver of the $q$-th V2V link on the $w$-th sub-band.
		The second is channel gain ${h_{wq}}\left[ w \right]$ originating from the transmitter of the $w$-th V2V link to the receiver of the $q$-th V2V link on the $w$-th sub-band.
		So the signal-to-interference-noise ratio (SINR) for the $w$-th V2I link and the $q$-th V2V link on the $w$-th sub-band are:
		\begin{equation}\label{eq2}
			\vspace*{-0.2\baselineskip}
			SINR_w^{VI}\left[ w \right] = \frac{{P_w^{VI}{h_{w,B}}\left[ w \right]}}{{{\sigma ^2} + \sum\limits_q {{b_{q}}\left[ w \right]P_q^{VV}\left[ w \right]{h_{q,B}}\left[ w \right]} }},
		\end{equation}
		\begin{equation}\label{eq3}
			\vspace*{-0.2\baselineskip}
			SINR_q^{VV}\left[ w \right] = \frac{{P_q^{VV}\left[ w \right]{h_q}\left[ w \right]}}{{{\sigma ^2} + I\left[ w \right]}},
		\end{equation}
		where $I\left[ w \right] = P_w^{VI}{h_{w,q}}\left[ w \right] + \sum\limits_{{q'} \ne q} {{b_{{q'},w}}\left[ w \right]P_{{q'}}^{VV}\left[ w \right]{h_{{q'},q}}\left[ w \right]}$.
		${P_w^{VI}}$ and ${P_q^{VV}}$ are transmit powers of
		the $w$-th V2I transmitter and the $q$-th V2V transmitter respectively over the $w$-th sub-band.
		${b_{q}\left[ w \right]}$ is binary spectrum allocation indicators. ${b_{q}\left[ w \right]} = 1$ represents that the $q$-th V2V link shares the $w$-th V2I link. Conversely, ${b_{q}\left[ w \right]} = 0$.
		The encoded semantic information, suitable for transmission, is sent from the vehicle's transmitter either to BS) via V2I link or directly to another vehicle via V2V link.
		The received semantic signal is represented as:
		\begin{equation}\label{eq4}
			\vspace*{-0.2\baselineskip}
			Y\left[ w \right] = H\left[ w \right]X\left[ w \right] + N,
		\end{equation}
		where $H\left[ w \right]$ represents the channel gains for the V2I and V2V links on the $w$ sub-band.
		$Y\left[ w \right]$ represents the signal received in the related channel.
		\subsubsection{Receiver Model}
		The receiver includes a channel decoder and a semantic decoder. 
		Upon receiving the encoded semantic information, the BS or the receiving vehicle uses the channel decoding and semantic decoding parts to decode it back into its original textual form. So the decoded signal can be represented as:
		\begin{equation}\label{eq5}
			\vspace*{-0.2\baselineskip}
			\hat S\left[ w \right] = se_\mu ^{ - 1}\left( {ch_\nu ^{ - 1}{\rm{ }}\left( {Y\left[ w \right]} \right)} \right),
		\end{equation}
		where $\hat S\left[ w \right]$ represents the recovered sentence, $se_\mu ^{ - 1}{\rm{ }}\left( \cdot \right)$ is the semantic decoder network with the parameter $\mu$, and $ch_\nu ^{ - 1}\left( \cdot \right)$ is the channel decoder network with the parameter $\nu$.
		Cross-entropy (CE) is used as the loss function to quantify the difference between $\hat S\left[ w \right]$ and the original sentence $S\left[ w \right]$ which can be expressed as:
		\begin{equation}\label{eq6}
		\vspace*{-0.3\baselineskip}
		\begin{array}{l}
			{L_{CE}}(S\left[ w \right],\hat S\left[ w \right];\alpha ,\beta ,\mu ,\nu ) = \sum\limits_l {q({s_l}\left[ w \right])} log({s_l}\left[ w \right]) 
			\\+ (1 - q({s_l}\left[ w \right]))log(1 - p({s_l}\left[ w \right])),
		\end{array}
		\end{equation}
		where $q({s_{l}}\left[ w \right])$ is the actual probability of the $l$-th word appearing in $S\left[ w \right]$, and $p\left( {{s_{l}}\left[ w \right]} \right)$ is the predicted probability of the $l$ sentence ${s_{q,l}}\left[ w \right]$  appearing in the generated sentence $\hat S\left[ w \right]$. By minimizing the CE loss, the network can learn the word distribution $q({s_{l}}\left[ w \right])$ in the original sentence $S\left[ w \right]$.
		The mutual information (MI) between the transmitted symbol $X\left[ w \right]$ and the received symbol $Y\left[ w \right]$ can be expressed as:
		\begin{equation}\label{eq7}
		 \begin{array}{l}
		 	I\left( {X\left[ w \right];Y\left[ w \right]} \right)\\
		 	= \sum\limits_{X\left[ w \right] \times Y\left[ w \right]} {p\left( {X\left[ w \right],Y\left[ w \right]} \right)} \log \frac{{p\left( {X\left[ w \right],Y\left[ w \right]} \right)}}{{p\left( {X\left[ w \right]} \right)p\left( {Y\left[ w \right]} \right)}}\\
		 	= {E_{p\left( {X\left[ w \right],Y\left[ w \right]} \right)}}\left[ {\log \frac{{p\left( {X\left[ w \right],Y\left[ w \right]} \right)}}{{p\left( {X\left[ w \right]} \right)p\left( {Y\left[ w \right]} \right)}}} \right]
		 \end{array}.
		\end{equation}
		The marginal probabilities of transmitting $X\left[ w \right]$ and receiving ${Y\left[ w \right]}$ are denoted as $p\left( {X\left[ w \right]} \right)$ and $p\left( {Y\left[ w \right]} \right)$, respectively. $p\left( {X\left[ w \right],Y\left[ w \right]} \right)$ represents the joint probability of $X\left[ w \right]$ and ${Y\left[ w \right]}$.
		Since MI is equivalent to the 
		kullback-leibler (KL) divergence between marginal probability and joint probability, it can be expressed as:
		 \begin{equation}\label{eq8}
		 \begin{array}{l}
		 	I\left( {X\left[ w \right];Y\left[ w \right]} \right)\\
		 	= {D_{KL}}\left( {p\left( {X\left[ w \right],Y\left[ w \right]} \right)\parallel p\left( {X\left[ w \right]} \right)p\left( {Y\left[ w \right]} \right)} \right).
		 \end{array}
		 \end{equation}
		According to the Theorem \cite{RN87}, The KL divergence can be expressed with dual representation:
		\begin{equation}\label{eq9}
		\begin{array}{l}
			{D_{KL}}\left( {p\left( {X\left[ w \right],Y\left[ w \right]} \right)\parallel p\left( {X\left[ w \right]} \right)p\left( {Y\left[ w \right]} \right)} \right)\\
			\ge {{\rm E}_{p\left( {X\left[ w \right],Y\left[ w \right]} \right)}}\left[ T \right] - log\left( {{{\rm E}_{p\left( {X\left[ w \right]} \right)p\left( {Y\left[ w \right]} \right)}}\left[ {{e^T}} \right]} \right).
		\end{array}
		\end{equation}
		According to \eqref{eq8} and \eqref{eq9}, the lower bound of MI can be obtained. Train the network ${\rm T}$ using unsupervised methods, where ${\rm T}$ can be approximated using a neural network. The expectation can be estimated through sampling, and with an increase in the sample size, the estimate converges to the true value. The corresponding loss function can be expressed as:
		\begin{equation}\label{eq10}
		I\left( {X\left[ w \right];Y\left[ w \right]} \right) \ge {L_{MI}}(X\left[ w \right],Y\left[ w \right];\alpha, \beta, T ).
		\end{equation}
		The loss function is used to train a neural network and gets its parameters $\alpha$, $\beta$ and ${\rm T}$.

		To evaluate the semantic communication performance of text transmission, we employ semantic similarity \cite{RN33} as a performance metric $\xi$, which can be expressed as:
		\begin{equation}\label{eq11}
		\xi  = \frac{{B\left( {S\left[ w \right]} \right)B\left( {\hat S\left[ w \right]} \right)}}{{\left\| {B\left( {S\left[ w \right]} \right)} \right\|\left\| {B\left( {\hat S\left[ w \right]} \right)} \right\|}},
		\end{equation}
		$B\left( \cdot \right)$ represents the bidirectional encoder representation of the Transformers (BERT) model \cite{RN88} for sentences, 
		where $0\le\xi\le 1$, $\xi  = 1$ indicates the highest similarity between the two sentences, $\xi  = 0$ implies no similarity between the two sentences.
		$\xi$ measures how closely two sentences convey the same meaning, regardless of their wording. For instance, if "The oranges are sweet and delicious" is received as "Sweet oranges are very fragrant", traditional methods might see this as a failure due to wording differences. However, semantic similarity assesses if the underlying meaning is preserved, indicating successful transmission if $\xi$ is high. This metric is crucial for evaluating semantic communication performance, reflecting how well the intended meaning is maintained.
		\vspace{-0.3cm}
		\subsection{Novel Metrics for Spectrum Sharing in High-Speed Mobile Vehicular Networks}
		Assuming the size of the text datasets is $\Upsilon$, where each sentence $S_i$ is represented as ${S_i} = \left[ {S_{i,1}, S_{i,2}, S_{i,3}, \ldots , S_{i,l}, \ldots , S_{i,{l_i}}} \right]$, where ${l_i}$ is the length of sentence ${S_i}$.
		The total text dataset size $\Upsilon$ is the sum of all sentences in the dataset, represented as $\Upsilon = \sum\limits_{i = 1}^\Upsilon {S_i}$. We assume the semantic information of sentence ${S_i}$ is ${I_i}$. Therefore, the average semantic information for each sentence is $I = \sum\limits_{i = 1}^\Upsilon  {{I_i}p\left( {{S_i}} \right)}$, where $p\left( {{S_i}} \right)$ represents the probability of sentence ${S_i}$ appearing in the text dataset. Similarly, the average length for each sentences is $L = \sum\limits_{i = 1}^\Upsilon  {{l_i}p\left( {{S_i}} \right)}$. 
		
		The HSR in high-speed mobile vehicular networks on the wireless channel can be expressed as:
		\begin{equation}\label{eq12}
		HSR = W\frac{I}{{{u_q}L}}\xi,
		\end{equation}
		where ${u_q}$ is the average number of semantic symbols used for each word, its unit is semantic unit (sut) \cite{RN92}, thus ${u_q}L$ is the semantic symbols of each sentence.
		$\frac{I}{{{u_q}L}}$ indicates the semantic information contained in each semantic symbol.
		$W$ corresponds to the number of semantic symbols transmitted per second. 
		Efficient utilization of the channel bandwidth means that the bandwidth occupied by the transmitted semantic symbols per second equals the available bandwidth.
		According to \cite{RN92}, the function is defined as $\xi = \Psi \left( {{u_q}, SINR\left[ w \right]} \right)$, meaning that semantic similarity depends on the number of semantic symbols ${u_q}$ and SINR. In the high-mobility vehicular environment, we can calculate the semantic similarity based on the current SINR and the chosen ${u_q}$. The value of ${\xi}$ increases with more ${u_q}$ and better $SINR$. Detailed semantic info and higher $SINR$ enhance similarity by reducing distortion.
		
		The unit of HSR is $suts/s$.
		HSR measures the rate at which semantic information is successfully transmitted over the wireless channel, reflecting the quality of communication.
		In high-speed vehicular environments, maintaining high communication quality is essential for applications such as safety alerts and real-time navigation. HSR ensures that critical information is transmitted accurately and quickly, thereby enhancing the reliability and effectiveness of the communication system.
				
		Therefore, the HSSE is expressed as:
		\begin{equation}\label{eq13}
		HSS{E} = \frac{{HSR}}{W} = \frac{I}{{{u_q}L}}{\xi}.
		\end{equation}
		Its unit is $suts/s/Hz$. This metric quantifies the efficiency of transmitting semantic information, measured in symbols, per unit of available bandwidth. It provides insights into how effectively the communication system utilizes the available frequency spectrum for semantic information transmission.
		HSSE is related to semantic similarity ${\xi}$ according to (\ref{eq31}), and ${\xi}$ is highly correlated with the SINR \cite{RN92}. Since SINR measures the power ratio between signal, noise, and interference, it is directly relevant to energy consumption. Therefore, HSSE provides a holistic view of both semantic communication efficiency and energy consumption. 
		This metric is vital for ensuring that the communication system makes the best use of the limited spectrum resources, which is particularly important in the context of IoV where spectrum scarcity is a common issue. By focusing on the semantic content of the transmitted data, HSSE helps to maximize the information density within the available spectrum, leading to more efficient spectrum utilization.
	    The value of $\frac{I}{L}$ depends on the type of source \cite{RN92}. For the source in this paper, it is a constant that does not affect resource optimization. Therefore, this term $\frac{I}{L}$ can be omitted during optimization.
	    \vspace{-0.3cm}
	    \subsection{Optimization Problem}
	    In high-speed vehicular environments, the need for rapid and accurate transmission of semantic information to support various applications. 
	    On the one hand, maximizing HSSE in V2I links ${SSE_q^{VI}}$ ensures the efficient utilization of available spectrum resources to transmit as much semantic data as possible within the allocated bandwidth. 
	    On the other hand, maximizing the SRS in V2V links $p_q^{VV}\left[ w \right]$ ensures the accurate and prompt exchange of critical information, such as road conditions.

	    The problem of optimizing semantic-aware spectrum sharing in vehicular networks can be recognized as the maximization problem of HSSE in V2I links and SRS in V2V links in the considered system under the constraints of channel allocation, transmission power and the number of transmitted semantic symbols. The problem can be expressed as:
		\begin{subequations}\label{P0}
			\begin{equation}\label{eq14a}
			{P_0}:\mathop {\max }\limits_{{b_q},{P_q},{u_q}} \sum\limits_{w = 1}^W {HSSE_{q,w}^{VI}}  + \sum\limits_{w = 1}^W {\sum\limits_{q = 1}^Q {{b_q}\left[ w \right]} } p_q^{VV}\left[ w \right]
			\end{equation}
			\begin{equation}\label{eq14b}
			s.t.{b_q}\left[ w \right] \in \left\{ {0,1} \right\},\forall q \in Q;\forall w \in W,
			\end{equation}
			\begin{equation}\label{eq14c}
			\sum\limits_{j = 1}^W {{b_q}\left[ j \right] \le 1} ,
			\end{equation}
			\begin{equation}\label{eq14d}
			\sum\limits_{i = 1}^Q {{b_i}\left[ w \right] \le 1} ,
			\end{equation}
			\begin{equation}\label{eq14e}
			{u_q} \in \left\{ {0,1, \cdots ,{u_q}_{\max }} \right\},
			\end{equation}
			\begin{equation}\label{eq14f}
			{\xi _q}\left[ w \right] \ge {\xi _{th}}.
			\end{equation}
		\end{subequations}
		Constraints (\ref{eq14b}), (\ref{eq14c}) and (\ref{eq14d}) represent the channel allocation. (\ref{eq14c}) states that each V2V link can only share one of the V2I links, (\ref{eq14d}) specifies that each V2I link can only be allocated to one V2V link, (\ref{eq14e}) constrains the average number of semantic symbols for each word within a certain range, and (\ref{eq14f}) constrains the minimum value of semantic similarity.
		The physical meaning of this constraint is to ensure that the semantic content of the transmitted message maintains a predefined level of accuracy, guaranteeing that the essential information conveyed by the message is both comprehensible and relevant.
		
		On the one hand, because of the high mobility of vehicles and the dynamic changes in network topology, the quality of network connectivity will be affected by multipath fading, signal interference, and channel attenuation. Additionally, V2V and V2I communications among vehicles require consideration of mutual interference between different communication links, adding complexity to the problem. Traditional methods and protocols often fail to meet communication requirements in such a highly dynamic environment.
		
		On the other hand, optimization goals include maximizing the HSSE of V2I links and SRS in V2V communications. Balancing these two goals involves finding the optimal trade-off, and optimization problems typically need to be solved within limited time frames.
		As a result, we try to solve this optimal spectrum sharing problem using DRL.
		
		Among various DRL algorithms, we choose the SAC DRL algorithm because SAC's enhanced exploration capability makes it easier to find better patterns under multimodal rewards \cite{RN89} , making it suitable for maximizing both HSSE and SRS in this paper.
		
		In summary, we propose a spectrum sharing approach using SAC DRL called SSS, specifically designed for the challenges posed by high-speed mobile vehicular networks in a semantic-aware environment. 
	
		\section{Proposed SSS Algorithm Approach}
		Our main goal is to devise an optimal spectrum sharing strategy based on SAC DRL. This strategy aims to maximize both the HSSE and SRS in V2V links.
		In the following sections, we will first formulate the SSS framework, outlining the states, actions, and rewards. We will then introduce the specifics of SSS algorithm.
		\subsection{SSS framework}
		The SSS framework is composed of states, actions, and rewards. Treating BS as an intelligent agent, it interacts with an unknown communication environment to gain experience. The training process is divided into multiple episodes, and each episode consists of several steps.

		At the current step $t$, BS, acting as the intelligent agent, observes the current state ${s_q}\left( t \right)$. 
		Using the policy $\pi$, the agent generates actions ${a_q}\left( t \right)$ based on the observed state ${s_q}\left( t \right)$ during each step.
		Then, the BS intelligent agent receives the current reward $r\left( t \right)$ and observes the transition from the current state ${s_q}\left( t \right)$ to the next state ${s_q}\left( {t{\rm{ + }}1} \right)$. Following this, we will have a detailed explanation of the components of the SSS framework, including states, actions, and rewards.
		\subsubsection{State}
		For the current time step $t$, the state of BS agent includes the channel gain of V2V link ${h_{q,t}}\left[ w \right]$, the interference channel from other V2V transmitters ${h_{{q'}q,t}}\left[ w \right]$, the interference channel from all V2I link transmitters to the V2V link receivers ${h_{wq,t}}\left[ w \right]$ and the interference channel from V2V link transmitter to the BS ${h_{qB,t}}\left[ w \right]$. These can be accurately estimated by BS agent in each step $t$.
		We assume that the broadcast delay from the BS to all vehicles is negligible.
		Consequently, the channel gains of these self-channels and interference channels are denoted as ${H_q}\left( t \right) = \left\{ {{h_{q,t}}\left[ w \right],{h_{{q'}q,t}}\left[ w \right],{h_{wq,t}}\left[ w \right],{h_{qB,t}}\left[ w \right]} \right\}$. 
		
		To compute semantic similarity, it is also essential to include the SINR received on V2V receivers and V2I receivers, along with the number of semantic symbols. This can be expressed as ${Z_q}\left( t \right) = \left\{ {SINR_{w,t}^{VI}\left[ w \right],SINR_{q,t}^{VV}\left[ w \right],u_{w,t}^{VI},u_{q,t}^{VV}} \right\}$. 
		Additionally, the state comprises the remaining V2V effective semantic information $S{D_q}$, the remaining time budget ${T_q}$, the value of the current training episode ${e_q}\left( t \right)$, and the parameters ${\tau _q}\left( t \right)$, which control the exploration strategy of the agent during training. 
		
		The components mentioned above collectively constitute the comprehensive state of the V2V agent. Therefore, the current state can be represented as:
		\begin{equation}\label{eq15}
		{s_q}\left( t \right) = \left( {{H_q}\left( t \right),{Z_q}\left( t \right),S{D_q},{T_q},{e_q}\left( t \right),{\tau _q}\left( t \right)} \right).
		\end{equation}
		\subsubsection{Action}
		The key actions include channel selection for V2V links, transmission power control, and the selection of the length of semantic symbols by the V2V transmitter vehicle in the shared V2I link. 
		The agent's actions can be represented as a vector containing multiple elements.
		Specifically, it can be represented as shown in (\ref{eq16}):
		\begin{equation}\label{eq16}
		{a_q}\left( t \right) = \left( {{b_q\left( t \right)},{P_q}\left( t \right),{u_q}\left( t \right)} \right),
		\end{equation}
		where ${b_q}\left( t \right)$ represents the channel selection, ${P_q}\left( t \right)$ represents the transmission power chosen by the V2V link transmitter vehicles, and ${u_q}\left( t \right)$ represents the average number of semantic symbols used for each word which means the length of semantic symbols chosen by the V2V transmitter vehicle in the shared V2I link.
		These actions' values are usually confined within the range $\left[ { - 1,0} \right]$, which has been selected for efficient learning and convergence.
		Subsequently, these values undergo a linear mapping process to obtain actual parameters or lengths.
		Defining this action space enables flexible decision-making in semantically-aware spectrum sharing within high-speed mobile IoV environments, aiming to maximize the HSSE and SRS over V2V links.
		\subsubsection{Reward}
		The reward has a dual objective. Maximizing the semantic spectral efficiency of V2I is crucial for enhancing communication performance, especially for supporting high data rate entertainment services in mobile applications and ensuring a more efficient and reliable communication network. The reward for this aspect is expressed as:
		\begin{equation}\label{eq17}
		{r_1}\left( t \right) = \sum\limits_{w = 1}^W {HSSE_{q,w}^{VI}} \left( t \right).
		\end{equation}
		Additionally, we aim to increase SRS in V2V links, ensuring timely transmission of semantic information, thus allowing the V2V links to handle reliable periodic safety information.
		The success probability of delivering effective semantic payloads within the time budget ${T_q}$ can be defined as the probability that, within the given time frame, the cumulative semantic payload successfully exceeds a threshold defined by $\frac{{S{D_q}}}{{{T_q}}}$.
		
		If there are remaining payloads, the reward is HSR of the V2V link; otherwise, if all payloads ${S{D_q}}$ have been delivered, the reward is fixed at a constant $\varpi$, which is selected to be greater than or equal to the maximum cumulative sum of individual link rates in the V2V network. Thus, the reward for the second part is:
		\begin{equation}\label{eq18}
		{r_2}\left( t \right) = \left\{ {\begin{array}{*{20}{l}}
				{\sum\limits_{w = 1}^W {{b_{q,w}}\left( t \right)HS{R_{q,w}}} \left( t \right),if\quad S{D_q} \ge 0}\\
				{\varpi \qquad\qquad\qquad\qquad\quad ,otherwise}
		\end{array}} \right..
		\end{equation}
		Based on (\ref{eq17}) and (\ref{eq18}), the overall reward can be expressed as:
		\begin{equation}\label{eq19}
		r\left( t \right) = \lambda {r_1}\left( t \right) + \left( {1 - \lambda } \right){r_2}\left( t \right),
		\end{equation}
		where weighting factor $\lambda$ adjusts the trade-off between optimizing the two reward components. A larger $\lambda$ value emphasizes maximizing V2I HSSE, while a smaller $\lambda$ value prioritizes SRS in V2V links.
		
		Thus the expected long-term discounted reward under policy $\pi \left( {{a_q}\left( t \right)\left| {{s_q}\left( t \right)} \right.} \right)$ is calculated as:
		\begin{equation}\label{eq20}
		J\left( {\pi \left( {{a_q}\left( t \right)|{s_q}\left( t \right)} \right)} \right){\rm{ = }}{\mathbb E}\left[ {\sum\limits_{t = 0}^T {{\gamma ^{t - 1}}r\left( t \right)}  + \varepsilon {\rm H}\left( {\pi \left( { \cdot |{s_q}\left( t \right)} \right)} \right)} \right], 
		\end{equation}
		where $\gamma$ is the discount factor, weighing the importance of future rewards. ${\rm{H}}\left( {\pi \left( { \cdot |{s_q}\left( t \right)} \right)} \right) = \left[ {\log \pi \left( { \cdot |{s_q}\left( t \right)} \right)} \right]$ represents the entropy of the policy $\pi \left( {{a_q}\left( t \right)\left| {{s_q}\left( t \right)} \right.} \right)$ at state ${{s_q}\left( t \right)}$. $\varepsilon$ represents the balancing weight between exploring feasible policies and maximizing rewards which can be dynamically adjusted based on the state.
		The optimal ${\varepsilon ^{\rm{*}}}$ at state ${{s_q}\left( t \right)}$ is denoted as:
		\begin{equation}\label{eq21}
		{\varepsilon ^{\rm{*}}}{\rm{ = }}\arg \mathop {\min }\limits_\alpha  {\mathbb E}\left[ { - {a_q}\left( t \right)\log {\pi ^*}\left( {{a_q}\left( t \right)|{s_q}\left( t \right)} \right) - \varepsilon \overline {\rm H} } \right],
		\end{equation}
		where ${\pi ^*}\left( {{a_q}\left( t \right)|{s_q}\left( t \right)} \right)$ is the optimal policy that maximizes the expected long-term discounted rewards under ${{s_q}\left( t \right)}$ and ${a_q}\left( t \right)$. ${\overline {\rm H} }$ is the expected policy entropy.
		\subsection{SSS Algorithm}
		We will describe the training phase to obtain the optimal policy and then introduce the testing phase to evaluate the performance under the selected optimal policy with the trained model.
		
		SAC comprises five key neural networks: a policy network, two soft Q networks, and two target soft Q networks. 
		The soft Q networks and target soft Q networks are used to compute the corresponding Q value and target Q value, respectively. 
		The policy network aims to maximize the Q value and obtain the optimal policy ${\pi ^*}\left( {{a_q}\left( t \right)|{s_q}\left( t \right)} \right)$ by generating the probability distribution of actions given a state.
		
		It is assumed that the parameters of the policy network  are represented by $\rho$, while the parameters of the two soft Q networks are denoted as ${\theta _1}$ and ${\theta _2}$, respectively. Additionally, the parameters of the two target soft Q networks are represented by  ${\bar \theta _1}$ and ${\bar \theta _2}$. 
		The replay buffer, denoted as $R$, plays a crucial role in SAC algorithm by storing past experiences. This enables the algorithm to randomly sample and learn from previous transitions, promoting stable and efficient training.
		
		The SSS algorithm regarding the training phase and is illustrated in Algorithm \ref{al1} and is described in detail as follows, including how it is deployed in practical IoV networks.
		%%%%%%%%%%%%%%%%%%%%%%%%%%%%%
		\begin{algorithm}
			\caption{The SSS training phase solving ${P_0}$}
			\label{al1}
			\KwIn{$\rho, {\theta _1}, {\theta _2}, {\bar \theta _1}, {\bar \theta _2}, \varepsilon$ }
			\KwOut{the optimal policy parameter ${\rho ^*}$}
			Randomly initialize the $\rho, {\theta _1}, {\theta _2}, \varepsilon$\;
			Initialize parameters ${\bar \theta _1} = {\theta _1}, {\bar \theta _2} = {\theta _2}$;
			
			Initialize replay experience buffer $R$;
			
			\For{episode from $1$ to $episode\_\max$ do}
			{
				Adjust Exploration Rate;
				
				Update the environment every 20 episodes;
				
				Receive observation state ${s_q}\left( t \right)$;
				
				\For{$t\_step$ from $1$ to $T_{\text{max}}$ do}
				{
					\If{$i\_step > {\rm{ }}explore\_steps$}
					{
						Generate the action ${{a_q}\left( t \right)}$ with state ${s_q}\left( t \right)$;
					}
					\Else
					{
						Generate the action ${{a_q}\left( t \right)}$ randomly;
					}
					Take action ${{a_q}\left( t \right)}$ : Obtain rewards $r\left( t \right)$ and new state ${s_q}\left( {t + 1} \right)$;
					
					Store tuple $\left( {{s_q}\left( t \right),{a_q}\left( t \right),r\left( t \right),{s_q}\left( {t + 1} \right)} \right)$ in replay buffer $R$;
					
					\If{replay buffer $R \ge threshold$}
					{
						Randomly sample a batch of $bs$ experiences from the replay buffer $R$;
						
						Normalize the reward $r\left( t \right)$ according to (\ref{eq24});
						
						Update entropy parameter $\varepsilon$;
						
						Caculate the loss to update Q networks according to (\ref{eq28}) and (\ref{eq29});
						
						Update the parameter ${\theta _1}$ and ${\theta _2}$ of the Q networks based on (\ref{eq30}) and (\ref{eq31});
						
						Caculate the loss to update policy network according to (\ref{eq32});
						
						Update the parameters $\rho$ of the policy network based on (\ref{eq33});
						
						\If{$iteration > iteration\_threhold$}
						{
							Update the parameters ${\bar \theta _1}$ and ${\bar \theta _2}$ of the target Q networks based on (\ref{eq34}) and (\ref{eq35});
						}
					}
				}
			}
		\end{algorithm}
		%%%%%%%%%%%%%%%%%%%%%%%%%%%%%
		\subsubsection{Training stage}
		Firstly, the cloud server initializes the parameters $\rho$, ${\theta _1}$, and ${\theta _2}$, along with the weight parameters $\varepsilon$ and the discounted factor $\gamma$. 
		Set ${\bar \theta _1}$ and ${\bar \theta _2}$ equal to ${\theta _1}$ and ${\theta _2}$, respectively.
		Initialize the mobile vehicle network in the urban environment, specifying the number and positions of vehicles, as well as the values for fast and slow channel fading.
		The algorithm conducts $episode\_\max$ training episodes.
		For each episode, BS receives the state observations from the vehicles and then transmits to the cloud server.
		In the first episode, the exploration rate is reduced for the initial 80\% of the training and then maintained at a constant level.
		Additionally, it is necessary to initialize the positions of vehicles in the city's vehicular mobility environment and update the environment every 20 episodes. 
		Vehicle movement in the environment is simulated by updating their positions and directions, with movement possible in four directions: up, down, left, and right. 
		When a vehicle moves up or down, its position is adjusted according to the positions of the left and right lanes. Near an intersection, there is a 40$\%$ chance it will change direction to the left if it will pass through the intersection next time step. Similarly, if the vehicle moves right or left, its position is adjusted based on the positions of the upper and lower lanes. Near an intersection, there is a 40$\%$ chance it will change direction to up if it will pass through the intersection next time step.
		For keeping the same number of vehicles in the considered scenarios, we assume that when a vehicle crosses the environment boundary, its position is adjusted to the opposite edge.
		
		The update of slow fading includes path loss and shadow fading.
		\paragraph{Path Loss}
		Calculate the horizontal, vertical distances, and the Euclidean distance between V2V and V2I, denoted as $d_{{\rm{hor}}}^j$, $d_{{\rm{ver}}}^j$, and ${d^j}$, respectively, where $j = \left\{ {{\rm{V2V,V2I}}} \right\}$. 
		In the following expressions, we simplify by omitting the subscripts and referring as $d_{\text{hor}}$, $d_{\text{ver}}$, and $d$, respectively.
		The path loss is determined using the line-of-sight (LOS) path loss model if $\min \left( {d_{\text{hor}}, d_{\text{ver}}} \right) < 7$; otherwise, the non-line-of-sight (NLOS) path loss model is utilized if $\min \left( {d_{\text{hor}}, d_{\text{ver}}} \right) \ge 7$. Hence, the path loss model is formulated as:
		\begin{equation}\label{eq22}
			path\_loss = \left\{ {\begin{array}{*{20}{l}}
					{P{L_{Los}}(d)\quad ,if\min \left( {{d_{hor}},{d_{ver}}} \right) < 7}\\
					{P{L_{NLos}}({d_{hor}},{d_{ver}})\quad ,otherwise}
			\end{array}} \right..
		\end{equation}
		\paragraph{Shadowing}
		The update of shadow fading is given by:
		\begin{equation}\label{eq23}
		sh{a_{new}} = \exp \left( { - \frac{{\Delta d}}{{{d_{dec}}}}} \right) \cdot sh{a_{old}} + \sqrt {1 - \exp \left( { - 2 \cdot \frac{{\Delta d}}{{{d_{dec}}}}} \right)}  \cdot X,
		\end{equation}
		where $sh{a_{new}}$ and $sh{a_{old}}$ is the shadow fading at the new and old position, respectively.
		$X \sim N(0,{3^2})$ , ${\Delta d}$ represents the distance between paths. ${{d_{dec}}}$ is the decorrelation distance used to describe the correlation of shadow fading.
		All the aforementioned models and parameters comply with the C-V2X 3GPP standard communication protocol, ensuring interoperability across various IoV platforms and devices \cite{GPP}.
		\begin{figure*}[htbp]
			\centering
			\includegraphics[width=0.75\textwidth]{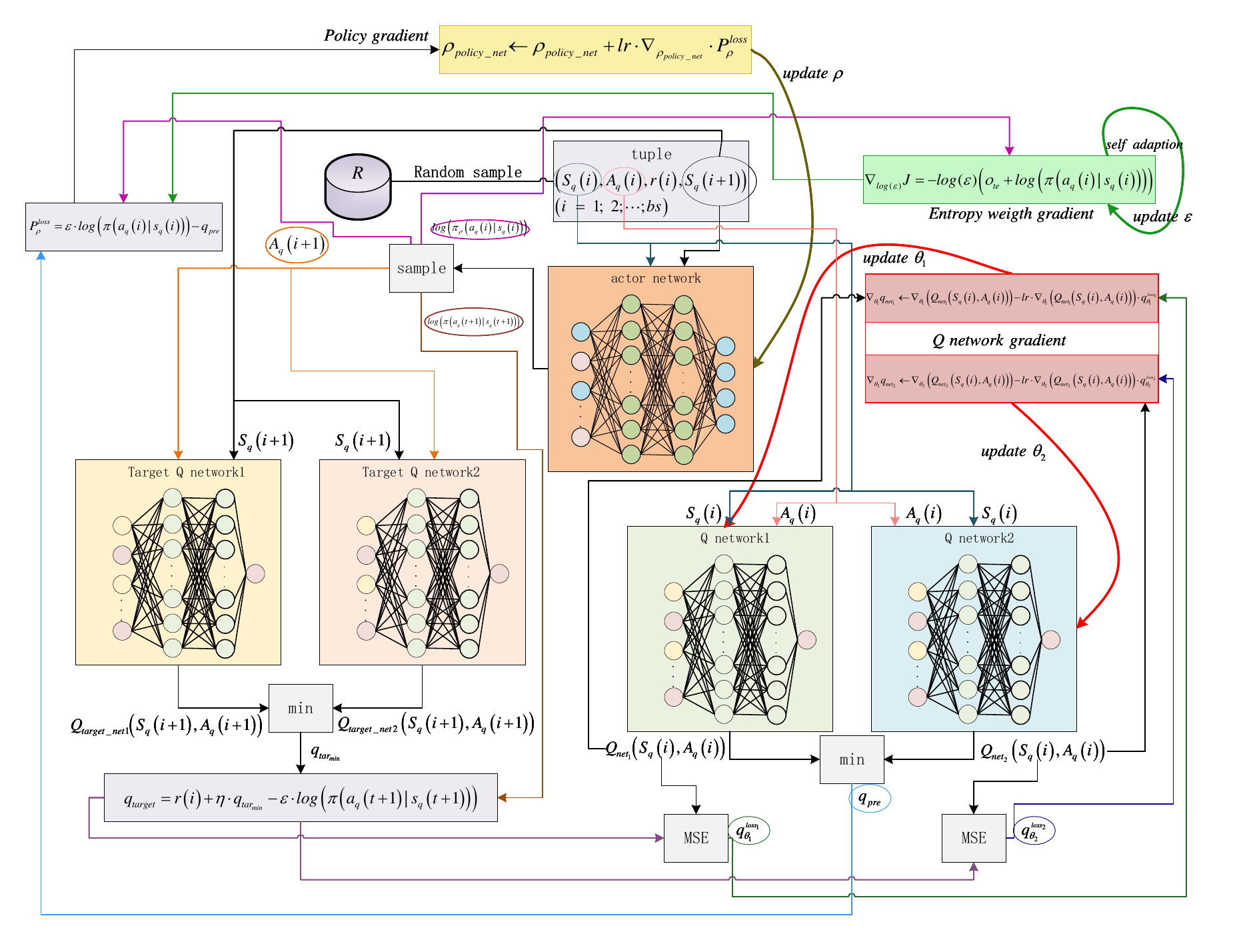}
			\caption{The process to update the parameters of networks}
			\label{fig4}
			\vspace{-0.2cm}
		\end{figure*}
		\paragraph{Fast fading}
		We represent the variations in channel gains using complex numbers drawn from a normal distribution with a mean of $0$ and a standard deviation of $1$.
		
		Next, initialize the size of the semantic data packets to be transmitted, the specified transmission time, and the link.
		The algorithm iterates for $T_{\text{max}}$ steps.
		For each time step, the cloud server receives the state observations from the vehicles.
		At $t=0$, ${s_q}\left( 0 \right)$ is obtained using (\ref{eq15}) and then is input into the policy network.  
		In the early stages of training, actions are randomly sampled from $U(-1, 1)$ to explore the environment without relying on the policy network output. Later, actions are determined by the mean and variance output of the policy network. 
		The action distribution follows a multidimensional Gaussian distribution and is scaled within $\left( { - 1,1} \right)$.
		Then, based on the action distribution ${\pi _\rho }\left( {\left. {{a_q}\left( 0 \right)} \right|{s_q}\left( 0 \right)} \right)$ with mean ${\mu _\rho }\left( 0 \right)$ and variance ${\sigma _\rho }\left( 0 \right)$, actions are sampled. 
		The BS agent takes action ${a_q}\left( 0 \right)$ and receives a reward $r\left( 0 \right)$ following (\ref{eq19}). Subsequently, the tuple $\left( {{s_q}\left( 0 \right),{a_q}\left( 0 \right),r\left( 0 \right),{s_q}\left( 1 \right)} \right)$ is stored in the replay buffer by the cloud sever to complete the iteration for $t=0$. This process can be iteratively repeated.
		The algorithm progresses to the next step at $t=1$. 
		Here, the state transition moves from ${s_q}\left( 0 \right)$ to ${s_q}\left( 1 \right)$. 
		If the number of samples in the experience replay buffer reaches a threshold,the server randomly selects a batch of $bs$ tuples from the replay buffer. Input the current state ${s_q}\left( i \right)$ and the next state ${s_q}\left( {i + 1} \right)$ into the policy network, one can obtain the current action ${a_q}\left( i \right)$, log probabilities of actions $log\left( {{\pi _\rho }\left( {{a_q}\left( i \right)|{s_q}\left( i \right)} \right)} \right)$ with its mean and log standard deviation, the next time step actions ${a_q}\left( {i + 1} \right)$ and the log probabilities of the next time step actions $log\left( {{\pi _\rho }\left( {{a_q}\left( {i + 1} \right)|{s_q}\left( {i + 1} \right)} \right)} \right)$.
		The actions $a_{_q}^{bs}$ in a batch $bs$ are normalized by the cloud server:
		\begin{equation}\label{eq24}
		a_{_q}^{bs} = rs \cdot \frac{{a_{_q}^{bs} - \mu \left( {a_{_q}^{bs}} \right)}}{{\sigma \left( {a_{_q}^{bs}} \right) + 1e - 6}},
		\end{equation}
		where $rs$ is the scaling parameter for normalization.
		$\mu \left( {a_{_q}^{bs}} \right)$ and $\sigma \left( {a_{_q}^{bs}} \right)$ are the mean and the standard deviation of the actions respectively.
		The loss can be expressed as:
		\begin{equation}\label{eq25}
			{\nabla _{\log(\varepsilon )}}J =  - \log(\varepsilon )\left( {o_{te} + \log\left( {\pi \left( {a_q(i)|s_q(i)} \right)} \right)} \right),
		\end{equation}
		where $o_{te}$ is the hyperparameter that controls the degree of policy exploration.
		Then, $\varepsilon$ is adapted, the value of the new entropy parameter is expressed as:
		\begin{equation}\label{eq26}
			\varepsilon  = \exp\left(\log(\varepsilon ) + \varepsilon_{optimizer} \cdot \nabla_{\log(\varepsilon )}J\right).
		\end{equation}
		Inputting the next state ${s_q}(i + 1)$
		and the next action ${a_q}(i + 1)$ into two target Q networks to calculate two target Q values $Q_{target\_net1}\left({s_q}(i + 1),{a_q}(i + 1)\right)$ and $Q_{target\_net2}\left({s_q}(i + 1),{a_q}(i + 1)\right)$, the smaller of the two values is used as $q_{tar_{\min}}$. Utilizing the Bellman equation to calculate $q_{target}$, the formula is expressed as:
		\begin{equation}\label{eq27}
			q_{target} = r(i) + \eta  \cdot q_{tar_{\min}},
		\end{equation}
		where $\eta$ is the discount factor.
		
		The current state ${s_q}(i)$ and action ${a_q}(i)$ are input into two Q networks to compute the predicted values ${q_{ne{t_1}}}$ and ${q_{ne{t_2}}}$. 
		The mean square error (MSE) is calculated between the predicted values and their corresponding target values ${q_{target}}$, yielding $q_{{\theta 1}}^{{los{s_1}}}$ and $q_{{\theta_2}}^{_{los{s_2}}}$, which can be expressed as:
		\begin{equation}\label{eq28}
			{q_{los{s_1}}} = {\left( {{Q_{ne{t_1}}}\left( {s_q(i),a_q(i)} \right) - {q_{target}}} \right)^2}.
		\end{equation}
		\begin{equation}\label{eq29}
			{q_{los{s_2}}} = {\left( {{Q_{ne{t_2}}}\left( {s_q(i),a_q(i)} \right) - {q_{target}}} \right)^2}.
		\end{equation}
		The Q network parameters ${\theta _1}$ and ${\theta _1}$ are updated using gradient descent (GD) to minimize ${q_{los{s_1}}}$ and ${q_{los{s_2}}}$, which can be expressed as:
		\begin{equation}\label{eq30}
			{\nabla _{{\theta _1}}}{q_{ne{t_1}}} \leftarrow {\nabla _{{\theta _1}}}\left( {{q_{ne{t_1}}}} \right) - lr \cdot {\nabla _{{\theta _1}}}\left( {{q_{ne{t_1}}}} \right) \cdot q_{{\theta _1}}^{_{los{s_1}}},
		\end{equation}
		\begin{equation}\label{eq31}
			{\nabla _{{\theta _2}}}{q_{ne{t_2}}} \leftarrow {\nabla _{{\theta _2}}}\left( {{q_{ne{t_2}}}} \right) - lr \cdot {\nabla _{{\theta _2}}}\left( {{q_{ne{t_2}}}} \right) \cdot q_{{\theta _2}}^{_{los{s_2}}},
		\end{equation}
		where $lr$ is the learning rate.
		The smaller of the predicted Q values ${q_{ne{t_1}}}$ and ${q_{ne{t_2}}}$ is chosen as ${q_{pre}}$.
		The policy is encouraged to generate actions with higher expected Q values by minimizing the cross-entropy loss $P_\rho ^{loss}$ between the policy probability and the expected Q values. It is expressed as:
		\begin{equation}\label{eq32}
			P_\rho ^{loss} = \varepsilon  \cdot \log\left( {\pi \left( {a_q(i)|s_q(i)} \right)} \right) - {q_{pre}}.
		\end{equation}
		The policy network parameters $\rho$ are updated using GD to maximize the expected Q values. The formula is expressed as:
		\begin{equation}\label{eq33}
			{\rho _{policy\_net}} \leftarrow {\rho _{policy\_net}} + lr \cdot {\nabla _{{\rho _{policy\_net}}}} \cdot P_\rho ^{loss}.
		\end{equation}
		Thus, we obtain three new parameters: the policy network parameter $\rho$, the soft Q networks parameters ${\theta _1}$ and ${\theta _2}$ as well as the weight parameters $\varepsilon$. 
		The update process for the current iteration is now complete. 
		The process to update the parameters of the two Q value networks and policy network is shown in Fig. \ref{fig4}.
		Then, move on to the next iteration, repeating this process until the iteration threshold is reached. Perform a soft update to update the parameters ${\bar \theta _1}$ and ${\bar \theta _2}$ of the two target Q networks. This process can be expressed as:
		\begin{equation}\label{eq34}
			{\bar \theta _1} \leftarrow (1 - \tau ) \cdot {\bar \theta _1} + \tau  \cdot {\theta _1},
		\end{equation}
		\begin{equation}\label{eq35}
			{\bar \theta _2} \leftarrow (1 - \tau ) \cdot {\bar \theta _2} + \tau  \cdot {\theta _2},
		\end{equation}
		$\tau$ is the soft update parameter, a fraction between 0 and 1 that controls the update speed of the target Q  network parameters.
		So, the above four parameters and two target networks ${\bar \theta _1}$ and ${\bar \theta _2}$ are updated to their latest values. 
		Then, the process continues to the next episode until $episode\_\max$ episodes are completed, obtaining the optimal parameters for the policy network ${\rho ^*}$.
		The Q networks and the policy network and their related parameters are all updated by the cloud sever.
		After training, the cloud server broadcasts the updated model parameters to the BS, which then disseminates them to all vehicles in the network.
		\begin{algorithm}
			\caption{The SSS testing phase}
			\label{al2}
			\KwIn{${\rho ^*}$ }
			\KwOut{the optimal channel allocation
				
				\qquad\qquad transmission power 
				
				\qquad\qquad the number of transmitted semantic symbols}
			Initialize environment parameters;
			
			\For{episode from $1$ to $episode\_test$ do}
			{
				Receive observation state ${s_q}\left( t \right)$;
				
				\For{$t\_step$ from $1$ to $T_{\text{max}}$ do}
				{
					Generate the action ${{a_q}\left( t \right)}$ according to the optimal policy ${\pi ^*}\left( {{a_q}\left( t \right)|{s_q}\left( t \right)} \right)$;
					
					Generate the responding  rewards $r\left( t \right)$ according to action ${{a_q}\left( t \right)}$;
					
					Obtain new state ${s_q}\left( {t + 1} \right)$;
				}
				
			}
		\end{algorithm}
		\subsubsection{Testing stage}
		During the testing phase, the first step involves loading a pre-trained SSS model, which is then utilized to execute tests and derive the optimal policy ${\pi ^*}\left( {{a_q}\left( t \right)|{s_q}\left( t \right)} \right)$ using optimized parameters ${\rho ^*}$.
		Similar to the training process, the initialization of the environment is required by the cloud server.
		During the testing phase, a loop is executed $episode_test$ times to obtain the optimal channel allocation, transmission power, and the number of transmitted semantic symbols. Performance metrics, such as HSSE and SRS in V2V links, are computed and recorded. The pseudo-code for the testing phase is provided in Algorithm \ref{al2}.

		\section{Simulation Results}
		In this section, We have evaluated the performance of the proposed SSS algorithm.
		\subsection{Simulation Settings and Dataset}
		We simulate a cellular-based vehicle communication network in an urban scenario shown in Fig. \ref{fig1} using Python 3.7 as the simulation tool, V2V and V2I communications adhere to an architecture with parameters set according to Annex A of 3GPP TR 36.885 \cite{RN90}. 
		The vehicle movement is modeled using a spatial Poisson process, and vehicle positions are updated every 100 ms in the simulation. 
		In urban settings, vehicles change their travel directions at intersections based on the following probability distribution: straight (50\%), left turn (25\%), and right turn (25\%).
		Each road consists of two lanes, totaling four lanes, with a lane width of 3.5 meters. The distance between adjacent intersections ${d_{{\rm{adj}}}}$ is 433 meters. 
		Throughout the simulation, all lanes maintain uniform vehicle density and speed, traveling at a constant speed of 36 kilometers per hour. 
		The minimum dimensions of the simulated area are 1299 meters by 750 meters. Detailed parameter settings are listed in Table~\ref{tab1}.
		\vspace{-0.3cm}
		\begin{table}[htbp]
		\begin{center}
			\caption{Parameters of System Model}
			\label{tab1}
			\begin{tabular}{|c|c|}
				\hline
				\textbf{Parameter} & \textbf{Value}\\
				\hline
				Numbers of $V2V$ links & 4\\
				\hline
				Numbers of $V2I$ links & 4\\
				\hline
				Numbers of road lanes& 4\\
				\hline
				Road lane width&3.5m\\
				\hline
				${d_{{\rm{adj}}}}$ & 433m \\
				\hline
				Simulated area&$1299{\rm{ }} \times 750{\rm{ }}{m^2}$\\
				\hline
				Absolute vehicle speed & $36km/h$\\
				\hline
				Carrier frequency${f_{carrier}}$&$2GHz$\\
				\hline
				Height of the BS ${h_{BS}}$ & $25m$ \\
				\hline
				Height of vehicle ${h_{vehicle}}$&$1.5m$\\
				\hline
				$V2V$ decorrelation distance & 10\\
				\hline
				$V2I$ decorrelation distance & 50\\
				\hline
				Position of the BS ${P_{BS}}$ & $(525.5,649.5)$\\
				\hline
				$V2V$ deviation of shadowing & 3$dB$\\
				\hline
				$V2I$ deviation of shadowing & 8$dB$\\
				\hline
				$V2I$ transmit power & 23$dBm$\\
				\hline
				$V2V$ transmit power & $\left[ { - 100,5,10,23} \right]$$dBm$\\
				\hline
				Noise power ${\sigma ^2}$& -114 $dBm$\\
				\hline
				Transform factor ${u_q}$& 5-40\\
				\hline
				$V2V$ semantic demand size & $\left[ {1,2, \cdots } \right] \times \frac{{1060}}{u_q}$$suts$\\
				\hline
				\end{tabular}
			\end{center}
		\end{table}
		\vspace{-0.2cm}
		The dataset selected for extracting semantic information from text is the European Parliament dataset \cite{RN91}, consisting of approximately 2.0 million sentences and 53 million words.
		The dataset is pre-processed, and sentence lengths are adjusted to range from 1 to 20 words. 
		The first 90\% of the dataset is used for training, and the remaining 10\% is for testing.
		The communication system for semantic information extraction illustrated in the Fig. \ref{fig2} . 
		The semantic encoding  utilizes a three-layer Transformer model with 128 units and 8 attention heads in each layer. 
		The channel encoding includes a Dense layer with 256 units, followed by a reshape layer to convert it into 16 units appropriate for the channel. The reshape layer, implemented as a dense layer, is then reshaped back to 256 units.
		The channel decoding comprises a Dense layer with 128 units, and the semantic decoding employs a three-layer Transformer decoder with 128 units and 8 attention heads in each layer. 
		EAll activation functions for the layers except those in the semantic encoding and decoding layers (which are linear) and the output prediction layer (which is Softmax) are ReLU.
		The input and output ends are both equipped with a MI estimation model, comprising three Dense layers with 256, 256, and 1 units, respectively, to output the MI result.
%		The specific neural network settings for extracting semantic information are detailed in Table~\ref{tab2}.
%		\begin{table}[htbp]
%			\small
%			\centering
%			\caption{Neural Network settings for Semantic Extraction \cite{RN33}.}
%			\label{tab2}
%			\begin{tabular}{|c|c|c|c|}
%				\hline
%				\textbf{Structure} & \textbf{Layer Name} & \textbf{Units} & \textbf{Activation} \\
%				\hline
%				\multirow{3}{*}{\textbf{Transmitter}}
%				& 3×Trans Encoder & 128 (8 heads) & Linear \\
%				\cline{2-4} % Adds a line between rows 2 and 3
%				& Dense & 256 & Relu \\
%				\cline{2-4} % Adds a line between rows 3 and 4
%				& Dense & 16 & Relu \\
%				\hline
%				\textbf{Channel} & AWGN & None & \\
%				\hline
%				\textbf{Receiver)}
%				& Dense & 256 & Relu \\
%				\cline{2-4} % Adds a line between rows 2 and 3
%				& Dense & 128 & Relu \\
%				\cline{2-4} % Adds a line between rows 2 and 3
%				& 3×Trans Decoder & 128 (8 heads) & Linear \\
%				\hline
%				\textbf{Pre Layer} & Dictionary Size & Softmax & \\
%				\hline
%				\textbf{MI Model}
%				& Dense & 256 & Relu \\
%				\cline{2-4} % Adds a line between rows 2 and 3
%				& Dense & 256 & Relu \\
%				\cline{2-4} % Adds a line between rows 2 and 3
%				& Dense & 1 & Relu \\
%				\hline
%			\end{tabular}
%		\end{table}

		The power control options are limited to four power levels: $[-100, 5, 10, 23] , \text{dBm}$.
		It's important to note that selecting $-100 , \text{dBm}$ effectively results in zero V2V transmission power. 
		Therefore, the action space dimensionality is $4 \times M \times 2$, where each action represents the power, selected spectrum sub-band, and the length of semantic symbols chosen for the V2V link over this V2I and shared V2I link.
		
		The settings for SAC parameters are shown in Table~\ref{tab3}.
		\vspace{-0.4cm}
		\begin{table}[htbp]
			\begin{center}
				\caption{:Parameters of SAC.}
				\label{tab3}
				\begin{tabular}{|c|c|}
					\hline
					\textbf{Parameter} & \textbf{Value}\\
					\hline
					learning rate for Q network $soft\_q\_lr$ & $3e-4$\\
					\hline
					learning rate for policy network $policy\_lr$ & $3e-4$\\
					\hline
					learning rate for $\varepsilon \_lr$ & $3e-4$\\
					\hline
					Replay buffer capacity &$1e6$\\
					\hline
					hidden layers Dimensionality & 256 \\
					\hline
					Reward discount factor $\gamma$&0.99\\
					\hline
					Rewards scaling factor $rs$& 10\\
					\hline
					soft update parameter$\tau$ &0.01\\
					\hline
				\end{tabular}
			\end{center}
		\end{table}
		\vspace{-0.4cm}
		\subsection{Performance Evaluation}
		To evaluate the performance of our proposed SSS algorithm denoting it as $SAC\_with\_sc$. We use HSSE in V2I link and SRS in V2V links as performance metrics. We compare our algorithm with the following baseline methods:
		\begin{itemize}
		\item Semantic-aware Spectrum Sharing Algorithm based on  DDPG DRL: This baseline utilizes a DDPG DRL algorithm for spectrum sharing considering semantic data. In our comparisons, it is referred to as $DDPG\_with\_sc$.
  
		\item Bits Spectrum Sharing Algorithm based on SAC DRL \cite{wuqiong1}: In this baseline, actions involve selecting V2V links and determining the transmission power for the selected link, without considering the selection of semantic symbol lengths for each vehicle. We denote the baseline as $SAC\_without\_sc$.
		
		\item Bits Spectrum Sharing Algorithm based on DDQN DRL \cite{RN49}:
		Similar to the second baseline, the difference lies in the utilization of DDQN DRL. We donate the baseline as $DDQN\_without\_sc$ .
		
		\item Bits Spectrum Sharing Algorithm based on Random selection \cite{RN47}:
		At each time step, BS acting as agents randomly selects shared V2V links and transmission power for the selected link. We donate the baseline as $rand\_without\_sc$.
		\end{itemize}
		
		In traditional communication, semantic information is transmitted using bits rather than symbols explicitly designed for semantics. While each bit can be loosely considered a semantic symbol, it generally carries less semantic information compared to symbols used in semantic transmission methods like DeepSC.
		The equivalent HSSE can be expressed as \cite{RN33}, similar to equation (\ref{eq13}) :
		\begin{equation}\label{eq36}
		HSSE' = {R_{q,w}}\frac{I}{{{u_q}L}}\xi,
		\end{equation}
		where ${u_q}$ is defined as the transform factor, similar to equation (\ref{eq13}), with units in bits/word. For example, if a word consists of 5 letters and ASCII code is used for encoding, then ${u_q}$ is 40 bits/word.
		We assume that ${\xi _i}=1$, indicating no bit errors in conventional communications.
		
		\begin{figure*}[htbp]
			\centering
			{\includegraphics[width=0.65\columnwidth]{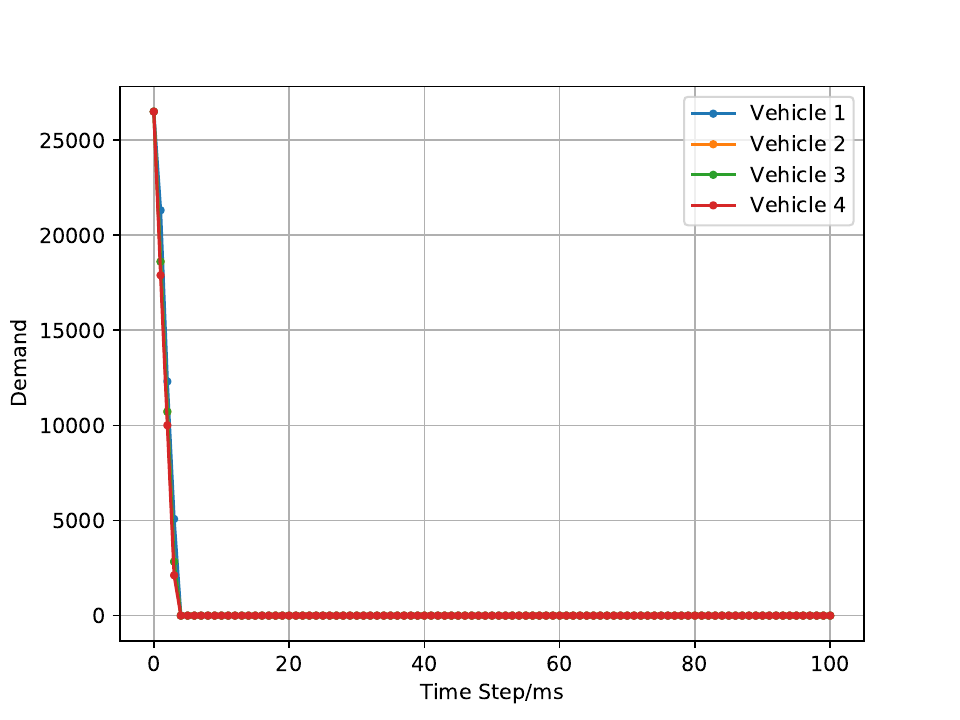}\label{fig5a}}
			{\includegraphics[width=0.65\columnwidth]{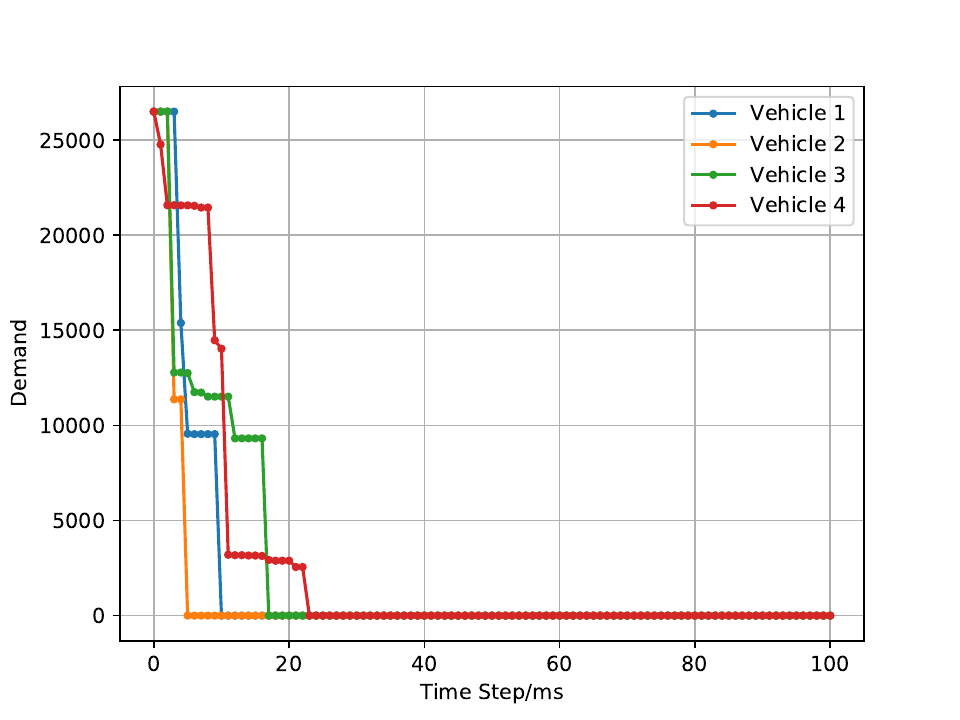}\label{fig5b}}
			{\includegraphics[width=0.65\columnwidth]{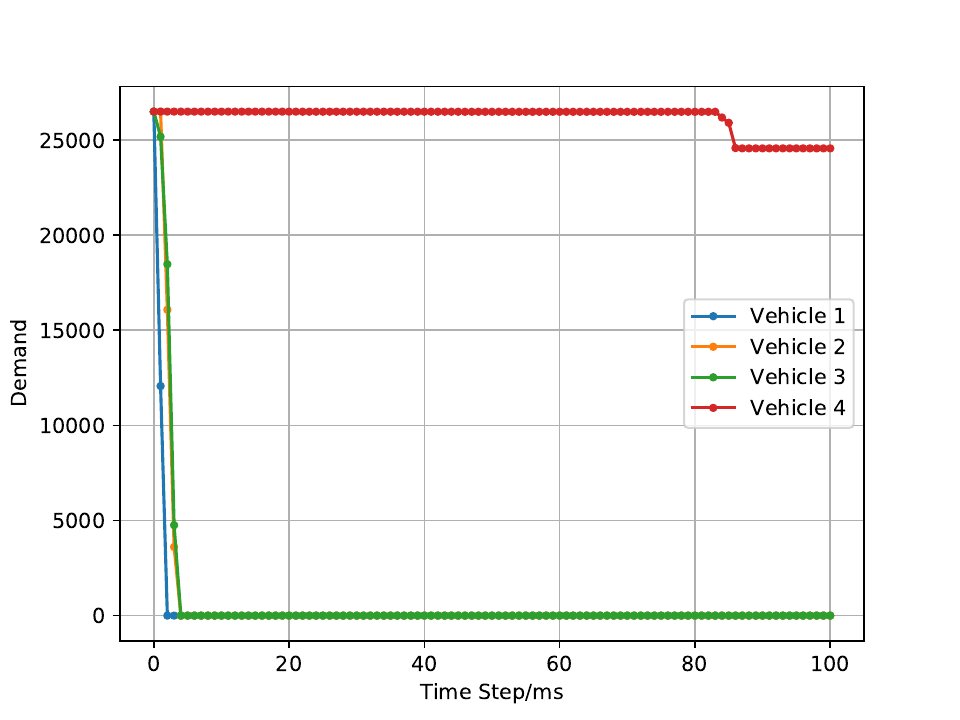}\label{fig5c}}
			\caption{The remaining demand of different algorithms.(a) SSS; (b) $DDQN\_without\_sc$; (c) $rand\_without\_sc$.}
			\label{fig5}
		\end{figure*}
		Fig.~\ref{fig5} illustrates the performance of three algorithms over time, measured in 100 milliseconds (ms). 
		The demand represents the successfully delivered payload by each algorithm. 
		Semantic information is converted to bits using a transforming factor ${u_q}$, resulting in a payload size of $25 \times 1060$ bits.
		We observe the success rates of the SSS algorithm and $DDQN\_without\_sc$ are 100\%, compared to 75\% with $rand\_without\_sc$ due to its lack of a learning mechanism to dynamically adjust decisions.
		The average response time of our SSS algorithm is 3.8 milliseconds, while $DDQN\_without\_sc$ algorithm has an average response time of 14 milliseconds, representing a 72.9\% reduction in latency.
		This improvement is primarily due to the SSS algorithm’s ability to leverage semantic information to deliver more valuable data and reduce redundancy, thus lowing transmission delay. Additionally, the algorithm's real-time decision-making capabilities and efficient resource allocation quickly adapt to changing conditions, ensuring low latency and high throughput.
		Low latency is also indicative of enhanced user experience, as it ensures timely and efficient communication, which is crucial for applications in IoV environments \cite{XIN4}.
		The successful payload delivery time varies for each vehicle in the $DDQN\_without\_sc$, as seen in the  Fig.~\ref{fig5}(a) and Fig.~\ref{fig5}(b).
	
		\vspace{-0.3cm}
		\begin{figure}[htbp]
			\centering
			\includegraphics[width=0.75\linewidth, scale=1.00]{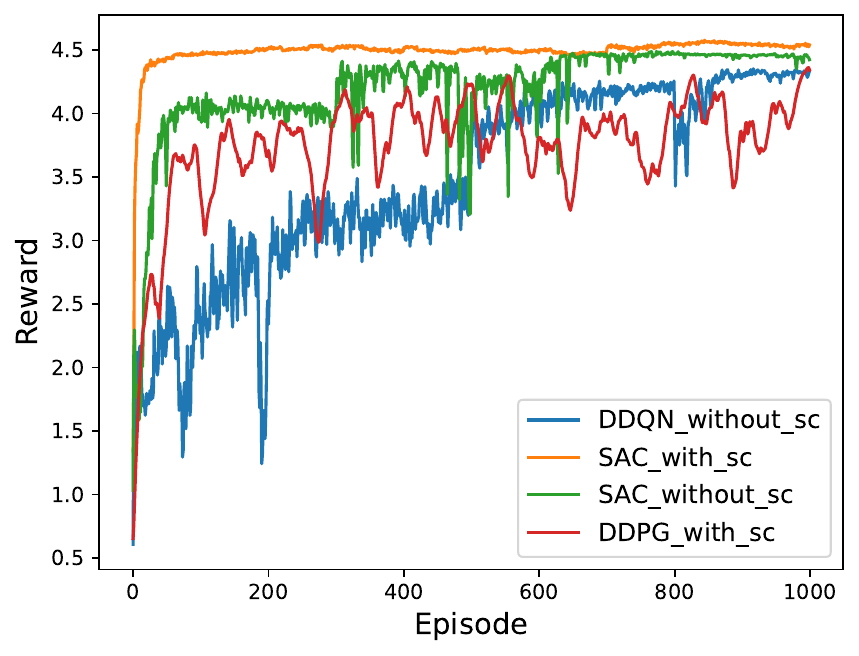}
			\vspace{-0.2cm}
			\caption{Training rewards across different spectrum sharing algorithms}
			\label{fig6}
			\vspace{-0.2cm}
		\end{figure}
		Fig. \ref{fig6} illustrates the reward evolution throughout the training process for both conventional reinforcement learning algorithms and our proposed SSS algorithm. 
		We can see that SSS algorithm exhibits greater stability in the task of semantic-aware spectrum sharing. In the early stages of training, our algorithm experiences a increase in rewards, maintaining stability in the later stages. 
		In comparison, $SAC\_without\_sc$ shows relatively stable rewards compared to the other two algorithms. However, it still lags behind our SSS algorithm. 
		This is attributed to the traditional approach lacking semantic awareness of the environment, and its action selection has limitations, resulting in limited information during decision-making.
		In $DDPG\_with\_sc$, the reward variability is unstable. 
		DDPG exhibits reward instability in complex semantic perception tasks due to the absence of the exploration-exploitation balance inherent in the SAC algorithm. 
		The rapid changes in semantic information in a highly dynamic vehicular network environment also contributes to the observed instability in rewards across different training phases.
		In contrast, $DDQN\_without\_sc$ exhibits a comparatively gradual increase in rewards.
		\vspace{-0.2cm}
		\begin{figure}[htbp]
			\centering
			\includegraphics[width=0.75\linewidth, scale=1.00]{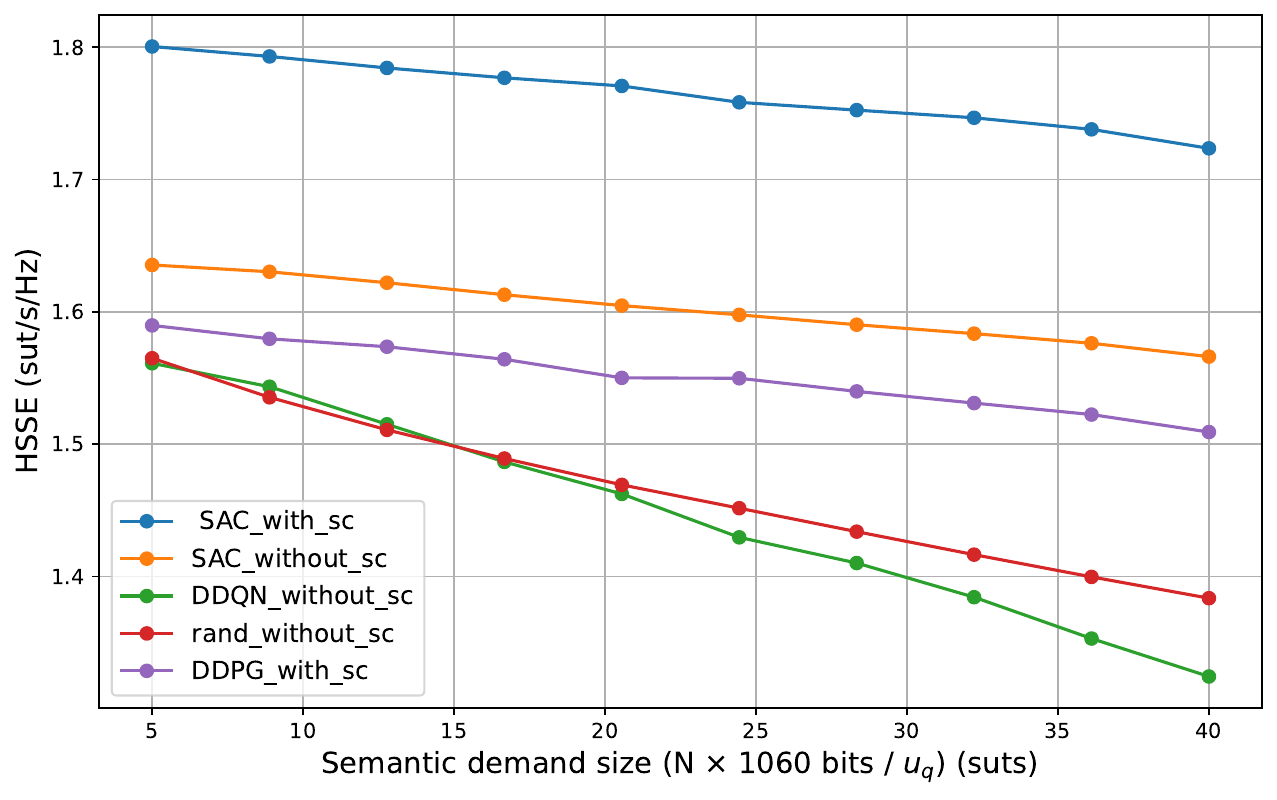}
			\vspace{-0.2cm}
			\caption{The HSSE versus semantic demand size}
			\label{fig7}
			\vspace{-0.2cm}
		\end{figure}
		Fig. \ref{fig7} shows how average HSSE varies with increasing semantic demand size.
		The transform factor ${u_q}$ is set to 20, corresponding to $V2V$ semantic demand sizes ranging from $\left[ {1,2, \cdots } \right] \times 530$ suts, with a transmission power of 23 $dBm$ for the V2I link transmitter.
		HSSE decreases with an increasing demand size, regardless of the algorithm used, which is because that increased semantic demand sizes lead to prolonged interference of V2V links with V2I, resulting in reduced HSSE and increased semantic demand sizes signifies a greater need for semantic information.
		Our SSS algorithm demonstrates notably higher HSSE which means the lower energy consumption and transmitting semantic information rather than bits. 
		Semantic information conveys more meaningful and user-oriented data, enabling our SSS algorithm to more effectively handle the transmission of semantic demand sizes, thereby reducing interference from V2V links to V2I links and enhancing HSSE.
		Furthermore, transmitting semantic information contains richer information than transmitting traditional bits at the same HSSE.
		Notably, $rand\_without\_sc$ performs better than outperforms $DDQN_without_sc$ when the semantic payload size is more than $20 \times 530$ suts.
		This is because in high-speed mobile IoV environments, $rand\_without\_sc$ exhibits a degree of flexibility, adapting better to dynamic environmental changes, while the $DDQN\_without\_sc$ may require more time to adapt to such variations.
		Furthermore, HSSE can also reflect user experience, as it indicates the efficiency and effectiveness of semantic information transmission.
		\vspace{-0.4cm}
		\begin{figure}[htbp]
			\centering
			\includegraphics[width=0.75\linewidth, scale=1.00]{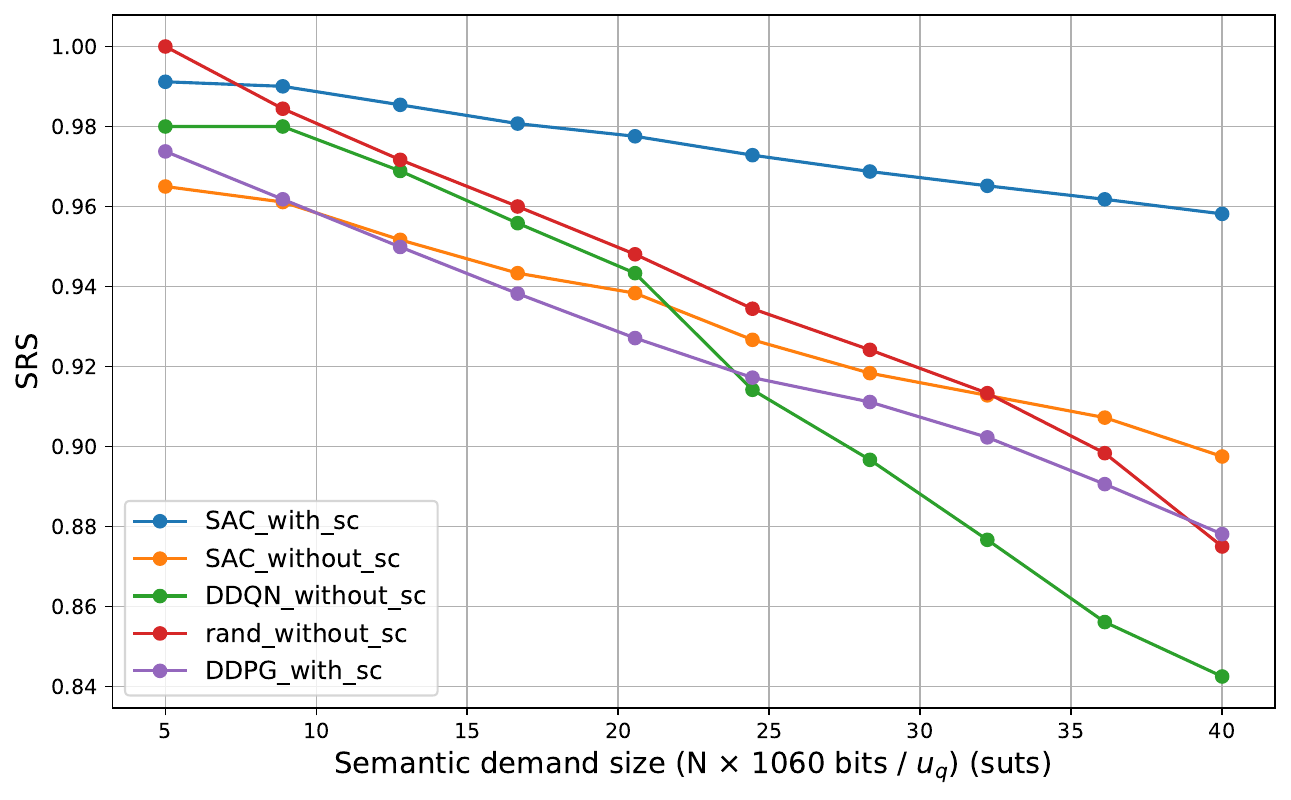}
			\vspace{-0.2cm}
			\caption{SRS oversus semantic demand size}
			\label{fig8}
			\vspace{-0.2cm}
		\end{figure}
		
		Fig. \ref{fig8} shows the relationship between semantic demand size and SRS.
		${u_q}$ is set as 20 and transmission power of the V2I link transmitter as 23 $dBm$.
		As semantic demand sizes increase, SRS decreases.
		This reduction occurs because a higher demand size requires more semantic information transmission within a given time, leading to a decline in SRS.
		Our SSS method exhibits a slower decline compared to the other algorithms which means the lower speed of energy consumption. We attribute to the use of compressed semantic information instead of traditional bits. The smaller size of semantic information reduces resource usage, maintaining higher V2V SRS.
		Beyond a semantic payload of $20 \times 530$ suts, the V2V transmission probability significantly decreases for $DDQN\_without\_sc$.
		This is because this method struggles to effectively adapt to the dynamic vehicular networks when dealing with a large amount of semantic payload. With increasing semantic payload, traditional bits transmission methods may lead to a sharp decline in transmission efficiency.
		In contrast, $rand\_without\_sc$ performs relatively well with smaller payloads due to its randomness but declines with larger payloads.
		Our proposed algorithm, by compressing and efficiently transmitting semantic information, demonstrates more stable and superior performance, effectively utilizing limited spectrum resources and improving SRS.
		
		\begin{figure}[htbp]
			\centering
			\includegraphics[width=0.75\linewidth, scale=1.00]{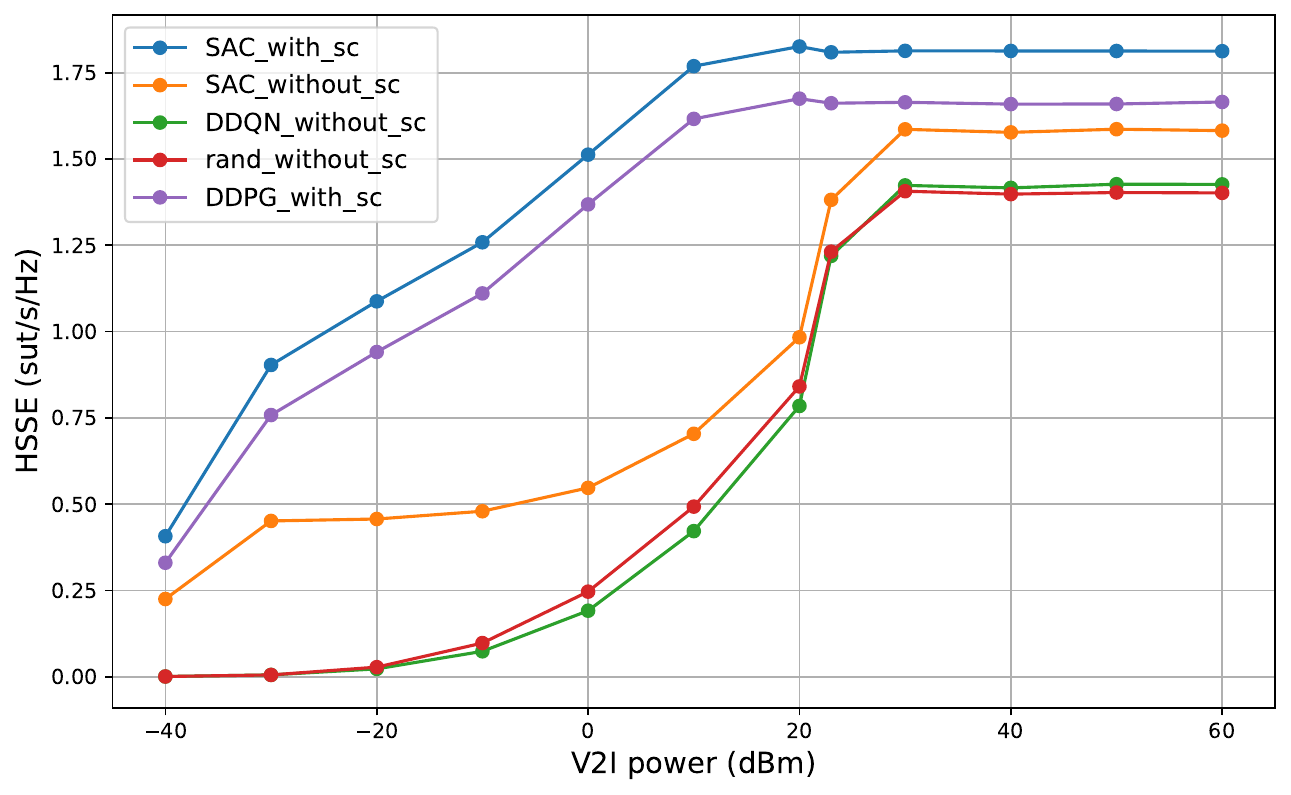}
			\vspace{-0.2cm}
			\caption{The HSSE versus V2I power}
			\label{fig9}
			\vspace{-0.2cm}
		\end{figure}
		Fig. \ref{fig9} shows the relationship between V2I link transmitter power and HSSE.
		${u_q}$ is set as 20 and semantic demand size as $25 \times 530{\rm{ }}$ suts.
		As V2I power increases, initially rises and then stabilizes, indicating an upper limit to HSSE with increasing power.
		Higher transmission power enables V2I link TO more effectively transmit semantic payloads, thereby enhancing HSSE. 
		However, beyond a certain point, further increases in power do not significantly impact HSSE, as the transmission of semantic information reaches a constant value.
		Our proposed algorithm consistently achieves higher HSSE which means the lower energy consumption across the entire power range compared to $DDQN\_without\_sc$, $rand\_without\_sc$, and $SAC\_without\_sc$, which transmit bits. Algorithms transmitting semantic information achieve higher HSSE.
		At low power levels, traditional algorithms exhibit a faster growth trend than algorithms transmitting semantic information.
		This is because traditional algorithms, focusing solely on the transmission of information rather than handling more complex semantic payloads, experience a rapid but still lower increase compared to our proposed algorithm.
		\vspace{-0.2cm}
		\begin{figure}[htbp]
			\centering
			\includegraphics[width=0.75\linewidth, scale=1.00]{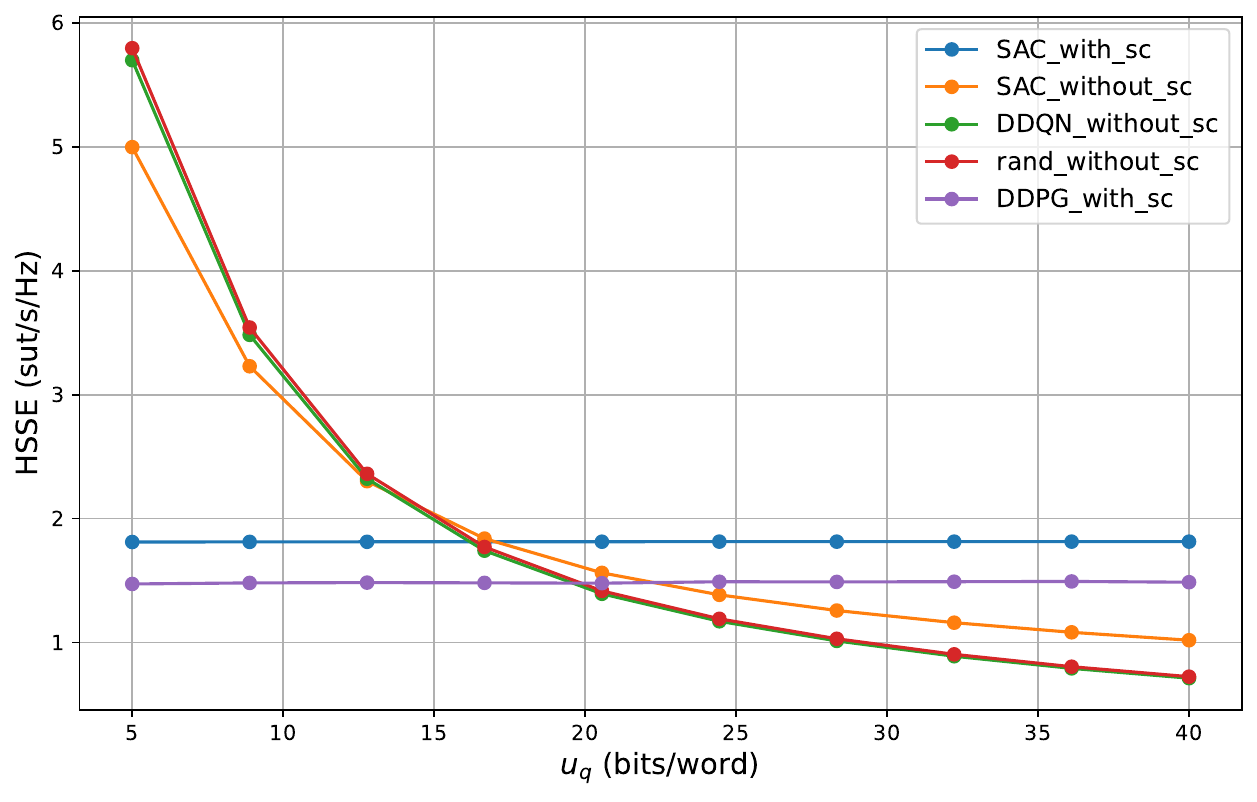}
			\caption{The HSSE versus transform factor ${u_q}$}
			\label{fig10}
			\vspace{-0.2cm}
		\end{figure}
		Fig. \ref{fig10} shows the relationship between ${u_q}$ and semantic spectrum efficiency.
		We set the semantic demand size as $25 \times 530{\rm{ }}$ suts and the transmission power of the V2I link transmitter to be 23 $dBm$.
		As ${u_q}$ changes, the HSSE of our proposed algorithm and $DDPG\_with\_sc$ remains constant. 
		This is because the system transmitting semantic information is independent of the transforming factor. Due to our definition of HSSE for traditional communication systems (according to (\ref{eq36}), which is the ratio of the transmission rate ${R_{q,w}}$ to the transforming factor ${u_q}$), traditional algorithms $DDQN\_without\_sc$, $rand\_without\_sc$, and $SAC\_without\_sc$ gradually decrease as ${u_q}$ increases. 
		Specifically, when ${u_q}$ is less than 17 $bits/word$, indicating a word encoding of less than 17 bits, our proposed algorithm performs better than traditional algorithms.
		This suggests that the choice of source coding scheme is crucial for traditional communication systems to achieve good performance through semantic information transmission.
		The primary advantage of using semantic information lies in its deeper understanding, beyond conventional bits. Semantic communication systems can more effectively compress information, achieving higher efficiency with the same spectrum resources. Therefore, despite traditional algorithms potentially demonstrating higher performance under certain conditions, semantic communication systems remain a more advanced and efficient choice.
	
		\vspace{-0.2cm}
		\begin{figure}[htbp]
			\centering
			\includegraphics[width=0.75\linewidth, scale=1.00]{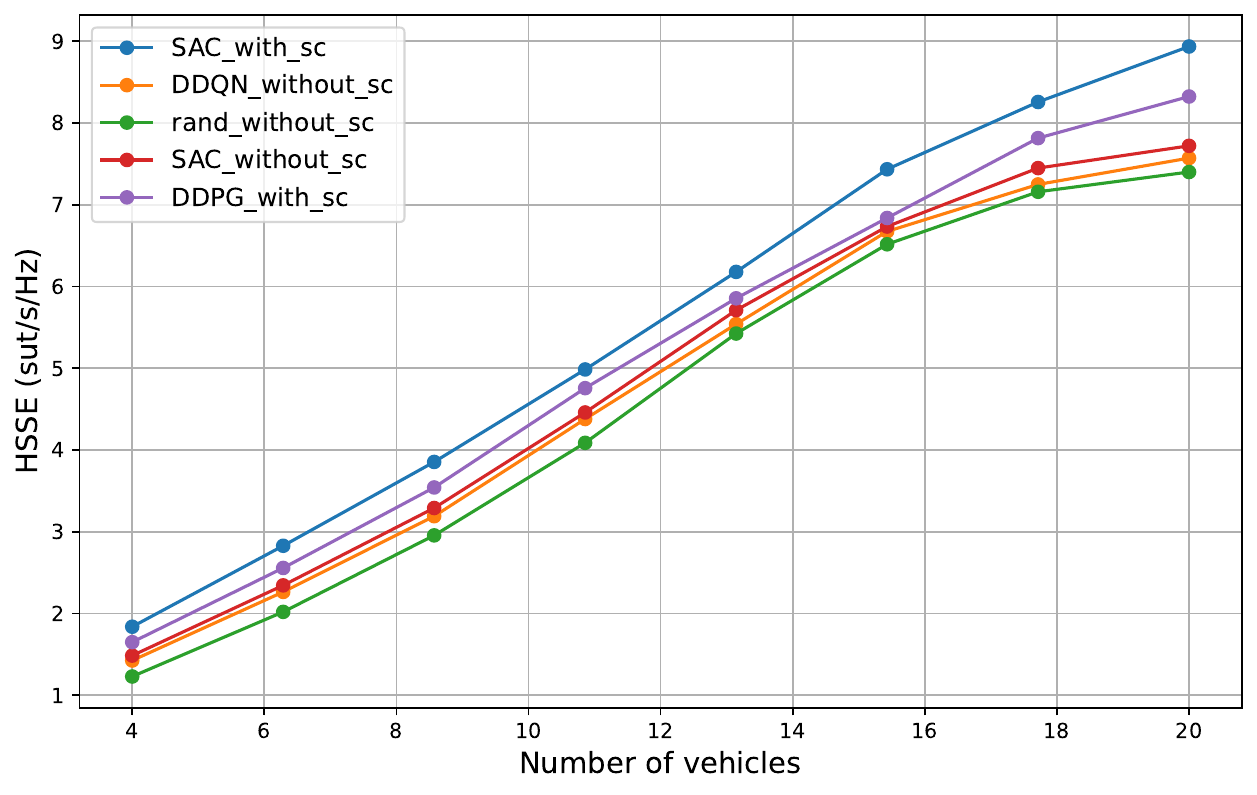}
			\caption{The HSSE versus vehicular number}
			\label{fig11}
			\vspace{-0.2cm}
		\end{figure}
		Fig. \ref{fig11} shows the relationship between the number of vehicles and HSSE, showing an overall increase as the vehicle count grows.
		Our proposed algorithm outperforms other $without\_sc$ methods. As the number of vehicles increases, more users compete for channel resources, causing HSSE to initially rise. However, as vehicles' number still growing, this also intensifies competition for limited spectrum resources, leading to potential bottlenecks in communication resources and spectrum availability. Our SSS excels in addressing this bottleneck by employing semantic information-based dynamic spectrum management.
		The DRL-based methods $without\_sc$ still outperform traditional random algorithms due to the optimized decision-making processes.
		\section{Conclusion}
%		In this paper, we addressed the challenges in high-speed vehicular networks by introducing semantic information into the spectrum sharing problem and proposed SSS, semantic-aware spectrum sharing scheme utilizing the SAC algorithm to maximize HSSE and the success probability of transmitting semantic data payloads. 
%		firstly, we introduced semantic information into the spectrum sharing problem of high-speed vehicular networks, proposing new communication metrics, HSR and HSSE. Then formulated a joint optimization problem aimed at maximizing HSSE and the transmission probabilities. Afterwards, we proposed a SSS algorithm to obtain optimal strategies. 
%		Simulation results have demonstrated that the SSS algorithm outperforms other baseline spectrum sharing algorithms.

		In this paper, we introduced semantic information into  high-speed mobile IoV environments, proposing the SSS scheme that utilizes the SAC DRL algorithm to maximize HSSE and SRS. 
		Simulation results illustrated the superior performance of the SSS algorithm to baseline spectrum sharing methods.
		The main conclusions are as follows:
		\begin{itemize}
			\item HSSE increases and stabilizes with higher V2I link transmission power, increases with a higher number of vehicles, but decreases with larger semantic demand sizes.
			\item The success probability of transmitting effective semantic data payloads decreases as semantic demand increases.
			\item When ${u_q}$ is more than 17 bits/word, indicating the need to transmit an average word length exceeding 17 bits in traditional communication systems, our semantic information transmission outperforms traditional bit-based systems. This highlights the efficiency and flexibility of our proposed algorithm.
		\end{itemize}
		
		In future work, we will further explore how to deploy and test our algorithm in real-world scenarios.

\nocite{*}
\bibliographystyle{IEEEtran}
\bibliography{ref}

 \end{document}